\documentclass[lettersize,journal]{IEEEtran}

\usepackage{amsmath,amsfonts}
\usepackage{algorithmic}
\usepackage{algorithm}
\usepackage{array}
\usepackage{textcomp}
\usepackage{stfloats}
\usepackage{url}
\usepackage{verbatim}
\usepackage{graphicx}
\usepackage{dsfont}

\usepackage[textsize=small]{todonotes}

\usepackage{cite}
\usepackage{iac_pkg}
\usepackage{eccvabbrv}
\usepackage{subcaption}
\usepackage{tikz}
\usetikzlibrary{decorations.pathreplacing} 
\usepackage{tablefootnote}
\usepackage{pifont}
\usepackage{amsmath,amssymb} 
\usepackage{color}
\usepackage{tikz-cd}
\usepackage{mathtools}
\usepackage{iac_pkg}
\usepackage{lipsum}
\usepackage{multirow}
\usepackage{tabularx, booktabs}
\usepackage{fontawesome5}
\usepackage{color}
\usepackage{wrapfig}
\usepackage{xspace}
\usepackage{overpic}
\usepackage{colortbl} 
\usepackage[export]{adjustbox}
\usepackage[textsize=small]{todonotes}
\usepackage{enumitem}
\definecolor{cvprblue}{rgb}{0.21,0.49,0.74}
\usepackage[breaklinks,colorlinks,allcolors=cvprblue,backref=false,pagebackref=false]{hyperref}
\usepackage{cleveref}

\definecolor{lightgray}{gray}{0.9}

\newtheorem{proposition}{\bfseries Proposition}

\definecolor{cornflowerblue}{rgb}{0.39, 0.58, 0.93}
\definecolor{citecolor}{RGB}{34,139,34}
\definecolor{lightgray}{RGB}{120, 120, 120}
\definecolor{lightred}{RGB}{235, 148, 149}
\definecolor{lightblue}{RGB}{173, 216, 229}
\definecolor{parula_1}{rgb}{0.2422,0.1504,0.6603}
\definecolor{plotBlue}{RGB}{31, 119, 180} 
\definecolor{plotOrange}{RGB}{255, 157, 38}  

\begin{document}

\title{Understanding Adversarial Training\\ with Energy-based Models}

\author{
Mujtaba Hussain Mirza,
Maria Rosaria Briglia,
Filippo Bartolucci,
Senad Beadini,
Giuseppe Lisanti,
Iacopo Masi

\thanks{This work was supported by
PNRR MUR PE0000013-FAIR under the MUR National
Recovery and Resilience Plan funded by the European
Union - NextGenerationEU and MUR PRIN 2022 project
20227YET9B ``AdVVent''.}
\thanks{Mujtaba Hussain Mirza, Maria Rosaria Briglia, and Iacopo Masi are with the OmnAI Lab research group at Sapienza University of Rome, Italy (mail: \{mirza, briglia, masi\}@di.uniroma1.it).}%
\thanks{Filippo Bartolucci and Giuseppe Lisanti are with the CVLab, University of Bologna, Italy (mail: \{filippo.bartolucci3, giuseppe.lisanti\}@unibo.it).}%
\thanks{Senad Beadini is with Eustema S.p.A., Italy (mail: s.beadini@eustema.it).}%
\thanks{}
}

\markboth{}%
{Mirza \MakeLowercase{\textit{et al.}}: Understanding Adversarial Training with Energy-based Models}


\maketitle

\begin{abstract}
We aim at using Energy-based Model (EBM) framework to better understand adversarial training (AT) in classifiers, and additionally to analyze the intrinsic generative capabilities of robust classifiers. By viewing standard classifiers through an energy lens, we begin by analyzing how the energies of adversarial examples, generated by various attacks, differ from those of the natural samples. The central focus of our work is to understand the critical phenomena of Catastrophic Overfitting (CO) and Robust Overfitting (RO) in AT from an energy perspective. We analyze the impact of existing AT approaches on the energy of samples during training and observe that the behavior of the ``delta energy'' —change in energy between original sample and its adversarial counterpart— diverges significantly when CO or RO occurs. After a thorough analysis of these energy dynamics and their relationship with overfitting, we propose a novel regularizer, the Delta Energy Regularizer (DER), designed to smoothen the energy landscape during training. We demonstrate that DER is effective in mitigating both CO and RO across multiple benchmarks. We further show that robust classifiers, when being used as generative models, have limits in handling trade-off between image quality and variability. We propose an improved technique based on a local class-wise principal component analysis (PCA) and energy-based guidance for better class-specific initialization and adaptive stopping, enhancing sample diversity and generation quality. Considering that we do not explicitly train for generative modeling, we achieve a competitive Inception Score (IS) and Fréchet inception distance (FID) compared to hybrid discriminative-generative models.
\end{abstract}
\section{Introduction}\label{sec:intro}

\begin{figure}[ht]
    \centering
    \begin{subfigure}[b]{1\linewidth}
        \centering
        \begin{overpic}[width=0.9\linewidth,height=3.2cm]{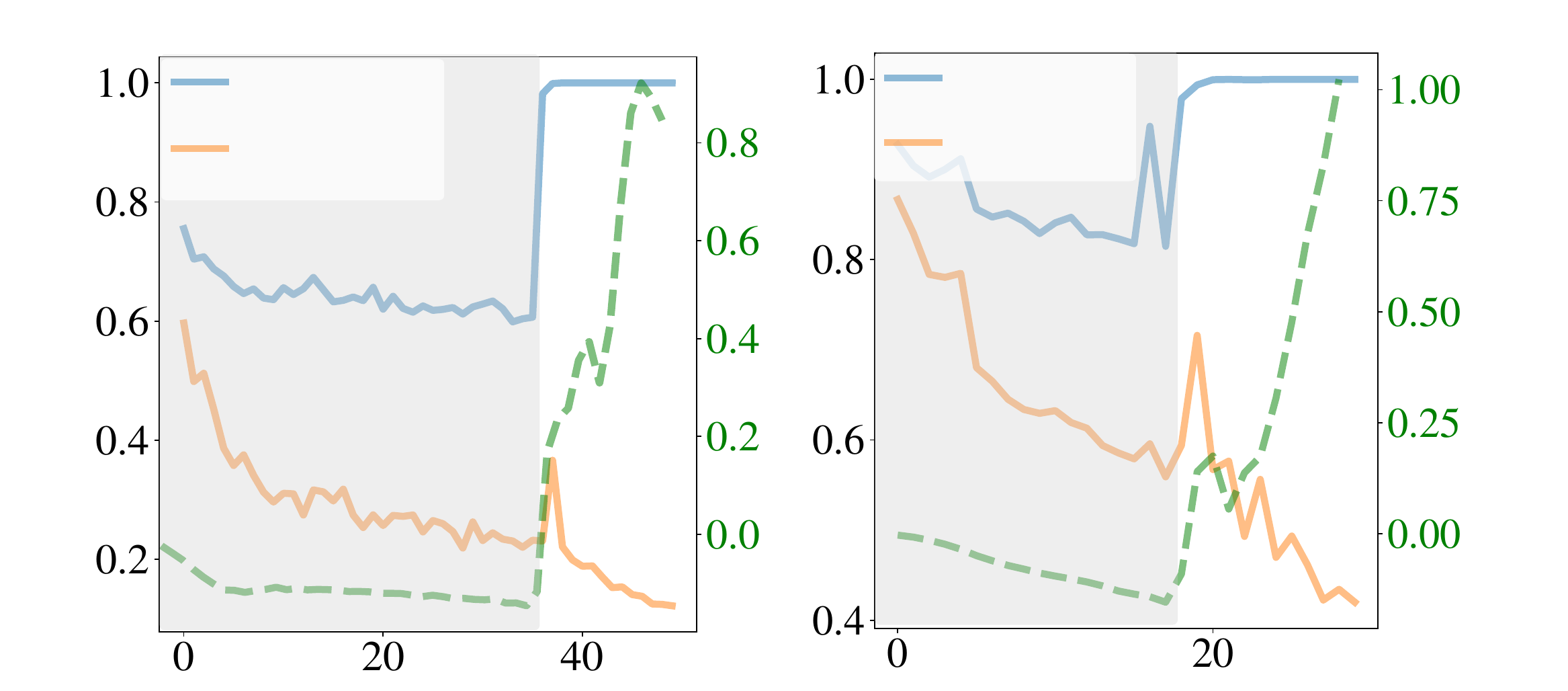}
            \put(24.5,45.5){\color{gray}Catastrophic Overfitting (CO)}
            \put(16.5,38.5){CIFAR-10}
            \put(15.2,34.5){\textcolor{black}{\tiny{PGD (test)}}}
            \put(15.2,30.8){\textcolor{black}{\tiny{Clean (train)}}}
            \put(1.5,10){\rotatebox{90}{\textcolor{black}{\small{Error Rate}}}}
            \put(97,15){\rotatebox{90}{\textcolor{citecolor}{\small{$\Delta\Ex$}}}}

            \put(62,38.7){CIFAR-100}
            \put(61,34.7){\textcolor{black}{\tiny{PGD (test)}}}
            \put(61,31){\textcolor{black}{\tiny{Clean (train)}}}
        \end{overpic}
        \caption{}\label{subfig:overfit-CO}
    \end{subfigure}
    \vfill
    \vspace{20pt}
    \begin{subfigure}[b]{1\linewidth}
        \centering
        \begin{overpic}[width=0.9\linewidth,height=3.2cm]{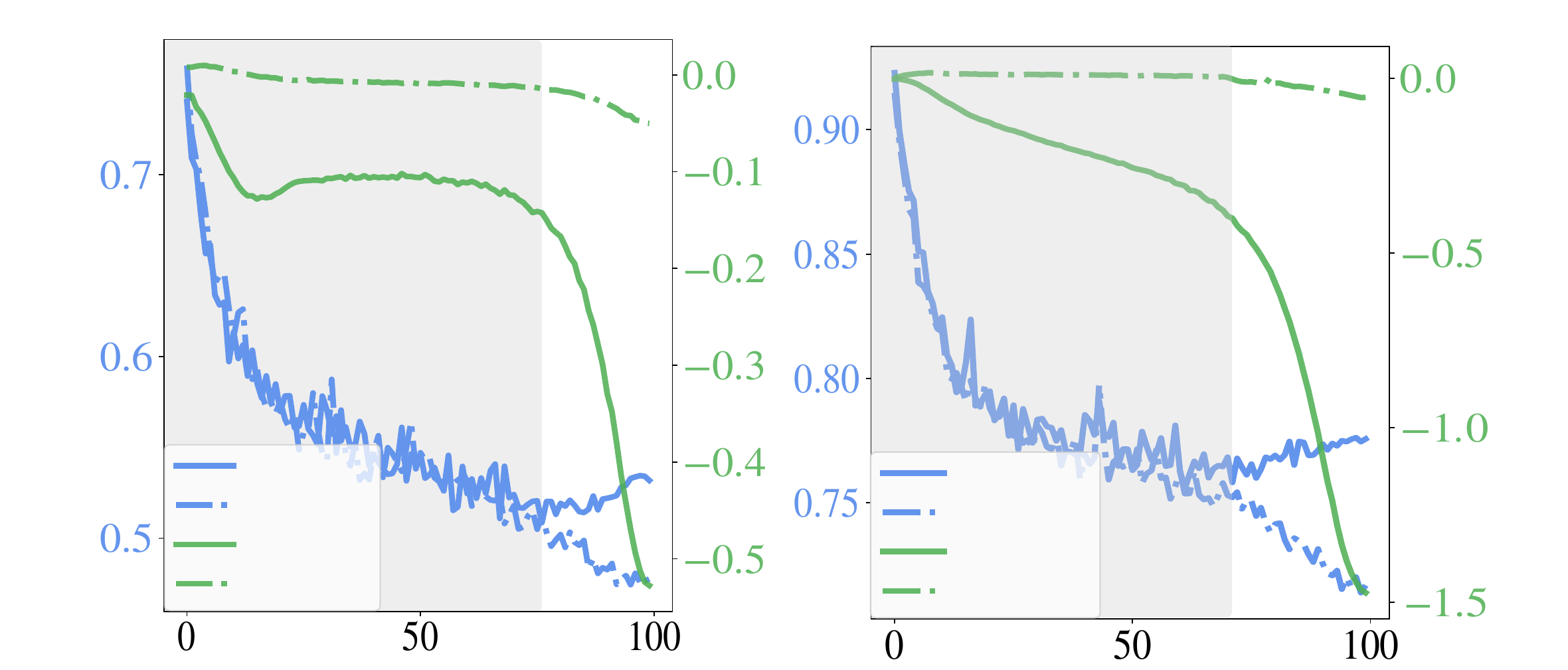}
             \put(26.5,44.5){\color{gray}Robust Overfitting (RO)}
             \put(16.5,38.75){CIFAR-10}
            \put(62,38.7){CIFAR-100}
            \put(15.2,11.4){\textcolor{black}{\tiny{SAT}}}
            \put(15.2,9.1){\textcolor{black}{\tiny{TRADES}}}
            \put(15.2,6.8){\textcolor{black}{\tiny{SAT}}}
            \put(15.2,4.5){\textcolor{black}{\tiny{TRADES}}}
            \put(61,11.1){\textcolor{black}{\tiny{SAT}}}
            \put(61,8.8){\textcolor{black}{\tiny{TRADES}}}
            \put(61,6.5){\textcolor{black}{\tiny{SAT}}}
            \put(61,4.2){\textcolor{black}{\tiny{TRADES}}}
            \put(1.5,10){\rotatebox{90}{\textcolor{cornflowerblue}{\small{Error Rate}}}}
            \put(97,15){\rotatebox{90}{\textcolor{citecolor}{\small{$\Delta\Ex$}}}}
            \put(23.5,-1.1){\textcolor{black}{\tiny{Epochs}}}
            \put(69.5,-1.5){\textcolor{black}{\tiny{Epochs}}}
        \end{overpic}
        \caption{}\label{subfig:overfit-RO}
    \end{subfigure}

    \caption{Analysis of $\Delta\Ex \doteq \Ex - \Exp$, computed on the training set (green, right axis), and error (PGD) on test data (blue, left axis) for CIFAR-10 (left column) and CIFAR-100 (right column) across catastrophic overfitting and robust overfitting. The onset of CO or RO is indicated by the start of the unshaded region.  
\textbf{(a)} For RS-FGSM, $\Delta\Ex$ increases sharply after CO, aligning with a sudden rise in test error, highlighting the impact of CO on energy dynamics.  
\textbf{(b)} Comparison of $\Delta\Ex$ in SAT (exhibits RO) and TRADES. In SAT, $\Delta E_{\net}$ decreases with rising test error, while in TRADES, it stays near zero as test error keeps decreasing.}

    \label{fig:overfit}
\end{figure}
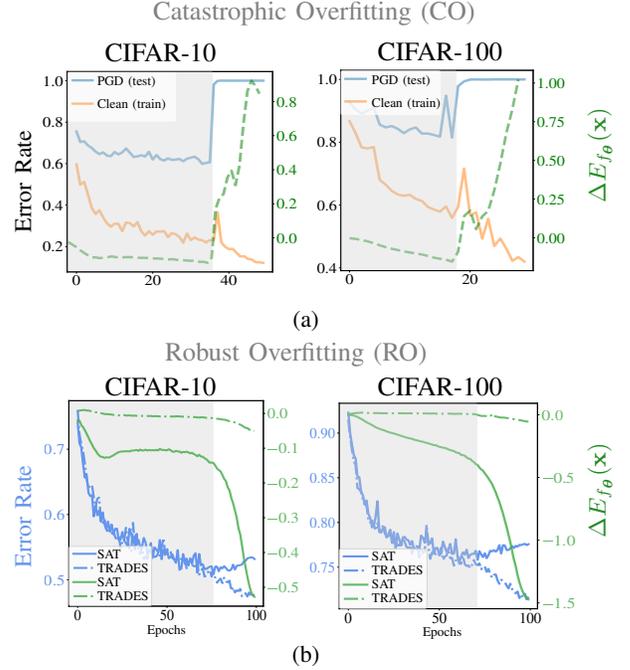

\IEEEPARstart{T}he pioneering paper by Szegedy \etal~\cite{szegedy2014} unveiled a striking revelation about neural networks: their susceptibility to adversarial attacks, where small input perturbations can lead to drastic changes in their predictions. 
This discovery ignited a surge of research aimed at strengthening model robustness against such perturbations, with adversarial training (AT) emerging as the dominant strategy~\cite{goodfellow2014explaining,madry2017towards,zhang2019theoretically,shafahi2019adversarial,wong2020f}. In recent years, many approaches have been developed, from multistep AT~\cite{madry2017towards,zhang2019theoretically,wang2019improving} for better defense, to single step AT~\cite{shafahi2019adversarial,wong2020f,sriramanan2021towards,sri2020guided} for faster training times. The use of additional data, whether real~\cite{carmon2019unlabeled} or synthetic~\cite{gowal2021improving,wang2023better}, has also proven effective. Furthermore, training strategies that incorporate sample re-weighting, such as MART~\cite{wang2019improving} and MAIL~\cite{liu2021probabilistic}, have shown promising results.
Researchers have increasingly focused on understanding and mitigating challenges that undermine the effectiveness of adversarial defenses, such as robust overfitting (RO)~\cite{rice2020overfitting, mirza2024shedding} and catastrophic overfitting (CO)~\cite{wong2020f,kim2021understanding}, evaluating the performance of the model on standard benchmarks like \texttt{RobustBench}~\cite{croce2021robustbench}.
However, despite these advances, overall progress has reached a plateau, with performance gains resulting mainly from larger datasets \cite{carmon2019unlabeled,wang2023better} or architectural innovations \cite{peng2023robust}.
Rare breakthroughs aside~\cite{wu2020adversarial}, AT has struggled to achieve significant increases in robustness. 
Interestingly, while much focus has been placed on performance, the underlying mechanics of AT remain less explored. Few studies attempt to explain the surprising capabilities of robust classifiers, such as their generative abilities and improved calibration. A notable exception is Zhu et al.~\cite{zhu2021towards}, who took the first steps in linking AT with Energy-based Models (EBMs) \cite{grathwohl2019your}, followed by our prior work~\cite{mirza2024shedding} paving the way for a deeper understanding of adversarial robustness.

\begin{figure*}[!t]
\centering
\begin{overpic}[keepaspectratio=true,width=\linewidth]{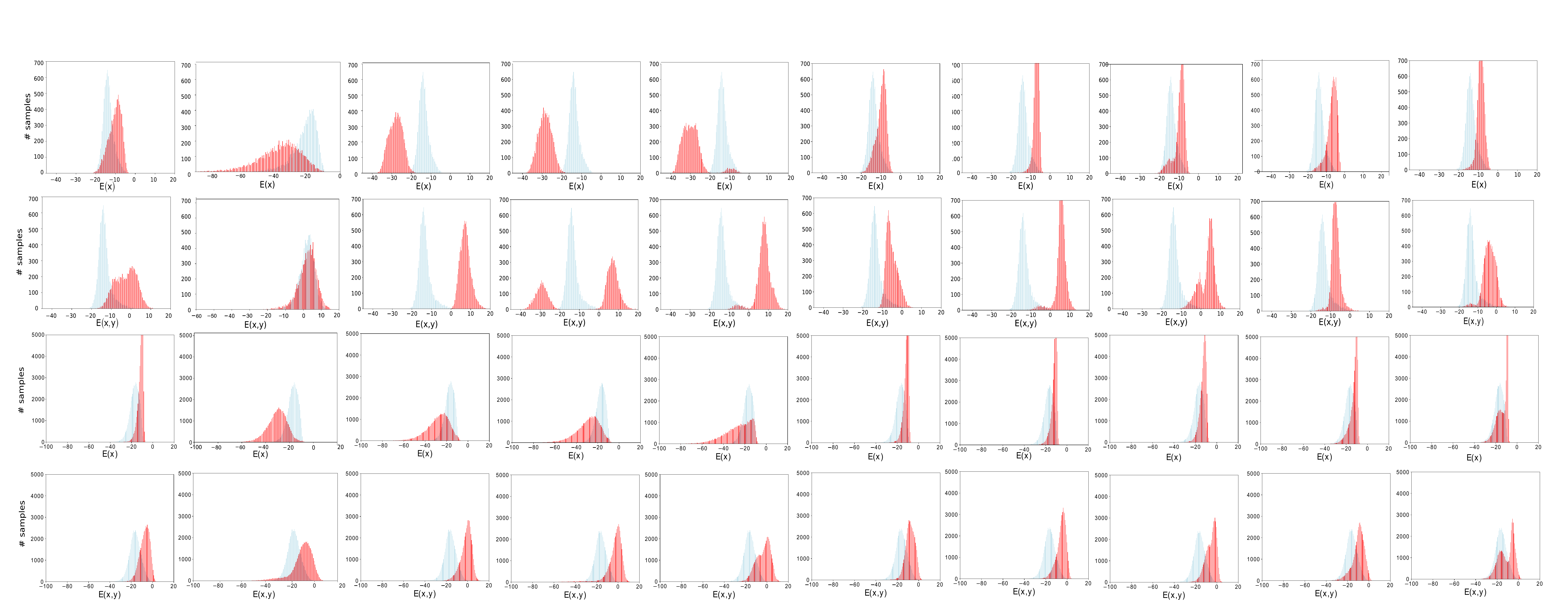}
    \put(-0.5,29){\rotatebox{90}{\scriptsize{$\Ex$}}}
    \put(-0.5,19){\rotatebox{90}{\scriptsize{$\Exy$}}}

    \put(-0.5,11){\rotatebox{90}{\scriptsize{$\Ex$}}}
    \put(-0.5,1.5){\rotatebox{90}{\scriptsize{$\Exy$}}}
    
     \put(99,30){\rotatebox{270}{{CIFAR-10}}}
        \put(99,14){\rotatebox{270}{{Imagenet}}}

    \put(4,36){\footnotesize{FGSM~\cite{goodfellow2014explaining}}}
    \put(13,36){\footnotesize{Diff-PGD~\cite{xue2023diffusion}}}
    \put(25,36){\footnotesize{PGD~\cite{madry2017towards}}}
    \put(33,36){\footnotesize{PGD-KL~\cite{zhang2019theoretically}}}
    \put(43,36){\footnotesize{APGD~\cite{croce2020reliable}}}
    \put(53,36){\footnotesize{CW~\cite{carlini2017towards}}}
    \put(60,36){\footnotesize{APGD-T~\cite{croce2020reliable}}}
    \put(69.5,36){\footnotesize{APGD-DLR~\cite{croce2020reliable}}}
    \put(81.5,36){\footnotesize{Square~\cite{andriushchenko2020square}}}
    \put(91,36){\footnotesize{CW-T~\cite{carlini2017towards}}}
    \end{overpic}
    \caption{Marginal — $\Ex$ — and joint — $\Exy$ — energy distributions for natural and adversarial inputs (with $y$ as the ground-truth label), shown for CIFAR-10 (rows 1–2) and ImageNet (rows 3–4) under various untargeted and targeted (-T) attacks. For this analysis, we use non-robust models trained on CIFAR-10 and ImageNet respectively. All attacks are generated with an input deformation constraint of $\ell_{\infty} \leq \epsilon = 8/255$. \textcolor{lightred}{\rule{0.4cm}{0.25cm}} indicates adversarial data, while \textcolor{lightblue}{\rule{0.4cm}{0.25cm}} indicates natural data.}
\label{fig:histogram-energy}
\end{figure*}
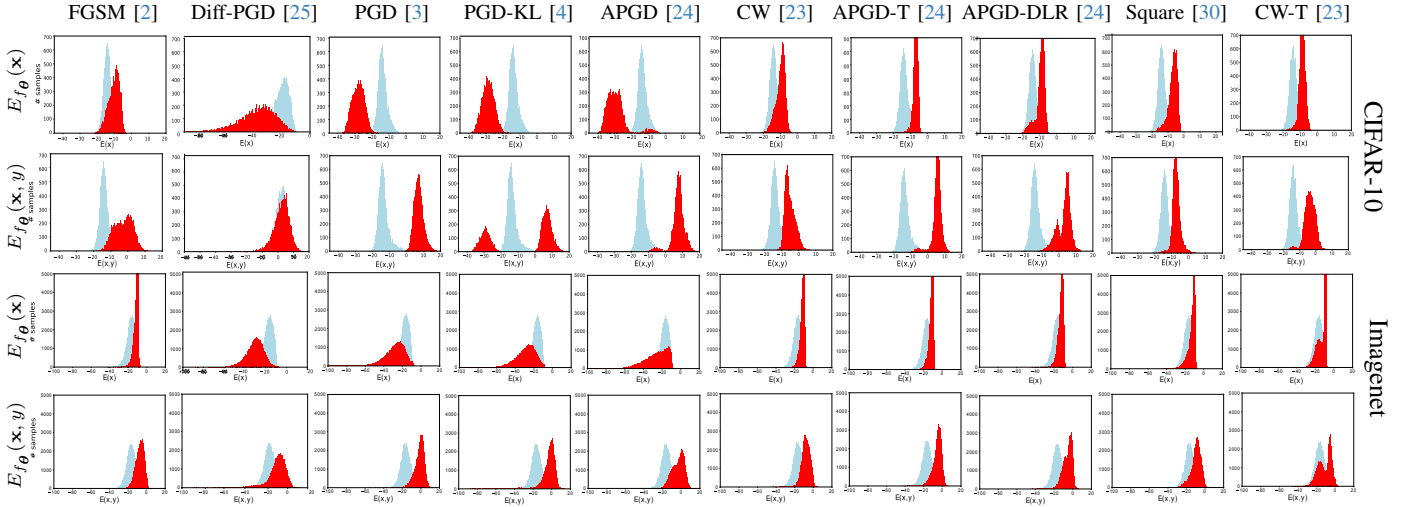

Adversarial attacks have been recognized as input points that cross the decision boundary---thus impacting $p_{\net}(y|\mbf{x})$---following~\cite{beadini2023exploring}, however, we illustrate a surprising yet strong correlation with $p_{\net}(\bx)$ for untargeted Projected Gradient Descent (PGD) attacks~\cite{madry2017towards}. Going beyond~\cite{beadini2023exploring}, we extend the analysis to a vast pool of attacks such as untargeted PGD~\cite{madry2017towards}, targeted attacks, CW~\cite{carlini2017towards}, TRADES (KL divergence)~\cite{zhang2019theoretically}, AutoAttack~\cite{croce2020reliable}, Fast Gradient Signed Method (FGSM)~\cite{goodfellow2014explaining}, diffusion-based adversarial samples such as Diff-PGD~\cite{xue2023diffusion} and show that different attacks induce difference shifts in the energy landscape on a \emph{non-robust} classifier, see \cref{fig:histogram-energy}.
However during AT, the adversarial samples generated typically exhibit energy levels similar to natural samples, but this changes noticeably when CO or RO occurs, particularly for specific subsets of samples. This is illustrated in \cref{fig:fig_quiver}, which visualizes the dynamics of changes in the joint energy, $E_{\net}(\bx,y)$, and marginal energy, $E_{\net}(\bx)$, of the samples when perturbed while training. We further monitor the delta energy $\Delta\Ex$ —the change in energy between original sample and its adversarial counterpart— for all training samples across training epochs, finding that these phenomena leave a distinct imprint on the overall energy dynamics, as indicated in \cref{fig:overfit}. Furthermore, we highlight that existing approaches that improve robustness, albeit unintentionally, reduce delta energy, indicating that smoother energy landscapes often accompany enhanced robustness. We find that TRADES \cite{zhang2019theoretically}, a state-of-the-art (SOTA) AT method, implicitly alleviates overfitting by means of aligning the natural energy with the adversarial one. While this method was not explicitly designed to address RO, our interpretation sheds light on may explain why it is less susceptible to RO than SAT~\cite{madry2017towards}\footnote{We refer to adversarial training (AT) as a generic procedure that regards all methods for robust classifiers (SAT, TRADES, MART, RS-FGSM) while SAT indicates Standard AT~\cite{madry2017towards}.}. Notably methods like TRADES \cite{zhang2019theoretically} and ALP \cite{kannan2018adversarial} have also been explored in single-step setting, but they do not effectively prevent CO, as demonstrated in \cite{lin2024eliminating}. Building on our insights, we propose training strategies for single-step and multi-step AT that incorporate a novel regularizer, the Delta Energy Regularizer (DER), designed to smoothen the energy landscape. Our main contributions are summarized as follows:

\begin{itemize}[itemsep=2pt, leftmargin=*]

\item Energy Landscape Analysis – We conduct a comprehensive analysis of the energy dynamics induced by adversarial training. By studying how the energies of natural samples change as they are perturbed within training to form adversarial examples, we reveal distinctive patterns associated with RO and CO. These insights deepen our understanding of these phenomena and enable us to mitigate them.
\item Analysis of Abnormal Adversarial Examples (AAEs) – We extend prior studies by investigating AAEs through the lens of energy. Our work analyzes them based on their energy behavior and demonstrates that, rather than the mere count of such examples, it is high-loss AAEs, those with high energy, that serve as critical indicators of CO. Furthermore, using the same framework, we provide an explanation for why CO does not occur in multi-step AT, showing that AAEs in this setting consistently exhibit lower energy.
\item We also offer experiments that demystify the role of misclassification~\cite{wang2019improving} and reconnect AT with energy and give a better explanation for the transferability of AT \wrt to training samples~\cite{losch2023onallexamples}. We theoretically show how rewriting TRADES as an EBM can better explain its capabilities.
\item Building on our energy analysis, we propose training strategies for single-step and multi-step AT that incorporate a novel regularizer, Delta Energy Regularizer (DER), designed to smoothen the energy landscape. We evaluate our method across varying adversarial budgets, attacks, and datasets, consistently demonstrating its effectiveness in preventing CO—even under larger perturbation bounds—while also avoiding RO to improve robustness in multi-step AT.
\item We further examine reweighting-based approaches and find that a part of the robustness gains from methods that weight the overall loss like MAIL-AT~\cite{liu2021probabilistic} and WEAT\cite{mirza2024shedding} can be attributed to scaling down the loss, rather than the reweighting strategy itself. Moreover, while methods such as MAIL-AT~\cite{liu2021probabilistic}, and $\text{DOM}_{RE}$~\cite{lin2024on} mitigate overfitting by down-weighting or removing overconfident samples—\emph{which we observe often exhibit low $\Delta \Ex$}—they tend to sacrifice robustness to Autoattack (AA). In contrast, DER smoothens the energy landscape around such samples to avoid RO, while also enhancing robustness against AA.

\end{itemize}

\section{Background and Prior Work}\label{sec:related}

\noindent Consider a set of labeled images 
$X = \{ (\bx,y) | \bx \in \mathbb{R}^{d}$ and $ y \in \{1,..,K\} \}$, assuming that each $(\bx,y)$ is generated from an underlying distribution $\mathcal{D}$; let $\net : \mathbb{R}^{d} \rightarrow \mathbb{R}^{K}$ be a classifier with parameters $\theta$ and  implemented with a DNN.

\subsection{Adversarial Robustness} 
\noindent Adversarial examples are inputs specifically crafted to deceive DNN by introducing a small perturbation. These perturbations, often imperceptible to humans, can cause the model to misclassify the input with high confidence. The existence of such vulnerabilities highlights fundamental weaknesses in standard training paradigms and has motivated extensive research on adversarial robustness.
Despite different proposed defense approaches, AT, which uses adversarial examples while training, remains the most effective empirical strategy. Formally, AT can be written as a min-max problem:
\begin{equation}
\min_{\theta} \mathbb{E}_{(\bx, y) \sim \mathcal{D}} \left[ \max_{\boldsymbol{\delta} \in \Delta} \mathcal{L}(f_\theta(\bx + \boldsymbol{\delta}), y) \right]
\label{eq:adversarial_training}
\end{equation}
\noindent where $\mathcal{L}$ is a loss function and $\Delta = \{ \pert \in  \mathbb{R}^{d}: ||\pert||_p \leq \epsilon \}$ is a set of feasible $\ell_p$ perturbations. The inner maximization seeks the worst-case perturbation \(\boldsymbol{\delta} \in \Delta\) to create an adversarial counterpart for $\bx$ denoted by $\bxa \doteq \bx+\pert \in \mathbb{R}^{d}$ in the input space by either increasing the loss in the output space (untargeted attack) or prompting a confident incorrect label (targeted attack) that minimizes the loss $\mathcal{L}$. The outer minimization aims to adjust the model parameters \(\theta\) to minimize the expected adversarial loss under the distribution \((\bx, y) \sim \mathcal{D}\). AT can be branched into single-step and multi-step, based on the number of gradient steps to approximate the inner maximization. 
\textit{Multi-step} approaches use an iterative method such as PGD~\cite{madry2017towards} to generate adversarial samples to provide a more accurate approximation of the worst-case perturbations. PGD can be formally defined as:
\[
\bxa_{t+1} = \mathbb{P}_\epsilon \Bigl( \bxa_t + \alpha \cdot \text{sign} \Bigl( \nabla_{\bxa_t} \mathcal{L}(\bxa, y; \theta) \Bigr) \Bigr)
\]
where the term \(\text{sign}(\cdot)\) denotes the sign function and \(\mathbb{P}_\epsilon\) is the projection operator, which projects into the surface of $\bx$'s neighbor $\epsilon$-ball, ensuring that \(\boldsymbol{\delta}\) remains within the allowed perturbation range. The initial step is \(\bxa_0 = \bx + \boldsymbol{\delta}_0\), where \(\boldsymbol{\delta}_0\) represents randomly initialized perturbations and \(\alpha\) is the step size for each attack iteration, \(\bxa_t\) denotes the adversarial example at iteration \(t\).
This methodology has attracted considerable interest and has received many variations. \cite{kannan2018adversarial} introduced an enhanced version of this defense using a
technique called logit pairing, encouraging logits to be similar for pairs of examples.
\cite{zhang2019theoretically} proposed TRADES, which leverages the Kullback-Leibler (KL) divergence to balance the trade-off between standard and robust accuracy, becoming the current default loss in several benchmarks.
\cite{cui2023decoupled} improved the KL divergence, addressing its asymmetry property, achieving remarkable results. 
Additionally, there are studies dedicated to exploring how DNN architecture impacts robustness~\cite{peng2023robust}.
\textit{Single-step} approaches use non-iterative methods such as the Fast Gradient Sign Method (FGSM)~\cite{goodfellow2014explaining} or a variant with an additional random step (R+FGSM)~\cite{tramèr2018ensemble} to generate adversarial samples. FGSM attack can be defined as follows:  
\[
\boldsymbol{\delta} = \epsilon \cdot \text{sign}\big(\nabla_{\bx} \mathcal{L}(\bx, y; \theta)\big)
\]
where \(\nabla_{\boldsymbol{x}} \mathcal{L}\) represents the backpropagated gradient of the loss \(\mathcal{L}\) with respect to the input \(\bx\). 
AT with this approach was previously considered a non-robust method, as models trained this way achieved nearly 0\% accuracy against stronger multi-step attacks like PGD.
However, recent research challenges this perception. 
Free adversarial training~\cite{shafahi2019adversarial} demonstrated that single-step methods could achieve strong performance by utilizing redundant batches and cumulative perturbations. Building on R+FGSM, \cite{wong2020f} proposed RS-FGSM, an approach that achieves performance comparable ~\cite{madry2017towards,shafahi2019adversarial}.

\subsection{Mitigation of overfitting in Adversarial Training}
\noindent 
CO is a phenomenon observed during training, more prevalently during single-step AT,  where robustness against multi-step attacks such as PGD suddenly decreases to 0\% just within a few epochs, whereas robustness against the FGSM attack rapidly increases. 
RO is a phenomenon mostly observed during multi-step AT, where the model's performance on adversarially perturbed training data continues to improve (i.e., the robust error decreases), but generalization ability to adversarial examples on unseen test data starts to degrade.
CO was first identified in \cite{wong2020f}, starting a line of work trying to explore and mitigate this problem. \cite{wong2020f} first suggested using early stopping to model training, however, implementing it requires continuous monitoring of robustness against PGD attacks throughout training, which is computationally expensive. Furthermore, \cite{vivek2020single} observed that single-step AT methods lead models to prevent the generation of single-step adversaries due to overfitting during the initial stages of training.
They empirically demonstrated that employing a dynamic dropout schedule can prevent this early overfitting to adversarial examples and make models robust to both single-step and multi-step attacks.
Noise-FGSM (N-FGSM)\cite{de2022make} finds that using a stronger noise around the clean samples, combined with not clipping around clean samples, is highly effective in avoiding CO for large perturbation radii. Other lines of work found that CO is closely related to anomalous gradient updates \cite{sriramanan2021towards,park2021reliably,GOLGOONI2023200258,andriushchenko2020understanding,li2022subspace,huang2023fast}. \cite{huang2023fast} shows that CO is instance-dependent and fitting instances with larger gradient norm is more likely to cause overfitting. Also, \cite{li2022subspace} identified a link between the gradient of each sample during AT and overfitting, also extending this insight to RO in multi-step AT. 
They propose a method that constrains AT to a carefully extracted subspace, effectively mitigating both RO and CO. 
\cite{sriramanan2021towards} improves single-step AT by introducing a regularizer to enforce local smoothness around the data samples. They also propose a two-step variant method that uses stable gradients from a weight-averaged model for better initialization and enhanced performance.
\cite{andriushchenko2020understanding} noted that RS-FGSM suffers from CO for larger perturbations and proposed GradAlign, a computationally expensive regularizer that enhances gradient alignment within the perturbation set to avoid CO.
On a more theoretical front, \cite{kim2021understanding} linked CO to distorted decision boundaries. \cite{lin2024eliminating} revisited the phenomenon, first noted by \cite{vivek2020single}, where certain adversarial examples generated during training exhibit lower loss than the original inputs. They termed these counterintuitive instances abnormal adversarial examples (AAEs). They proposed a method that explicitly prevents the generation of AAEs, achieving both efficiency and robustness by stabilizing decision boundaries.
RO was first investigated by \cite{schmidt2018adversarially} and argued for the need for a substantially large dataset for robust generalization. Later on, \cite{rice2020overfitting} extended the study, discovering that RO is a much more difficult problem than standard training, yet the culprit for it is still unknown. 
The work from~\cite{schmidt2018adversarially} has been extended in subsequent years by showing empirical evidence that larger datasets are even more essential for robust models: \cite{carmon2019unlabeled,alayrac2019labels,zhai2019adversarially} poured additional natural data from a similar distribution with pseudo labels. \cite{gowal2021improving} illustrated that training with images synthesized from generative models leads to an improvement in robustness. \cite{wang2023better} demonstrated that the use of synthesized images from more advanced generative models, such as diffusion models~\cite{ho2020denoising}, leads to superior adversarial robustness, setting a new SOTA in robust accuracy. 
Recently, \cite{dong2021exploring,lin2024on} hypothesizes that overfitting is due to difficult samples (hard to fit) that are closer to the decision boundary and the network ends up memorizing instead of learning. 
\cite{huang2023enhancing} optimizes the AT trajectories considering their dynamics, while others~\cite{chen2022sparsity,stutz2021relating,singla2021low,dong2021exploring,wu2020adversarial,chen2020robust} link generalization in AT to the flatness of the loss landscape. Other works like AWP~\cite{wu2020adversarial} also adversarially perturb model weights to flatten the weight landscape for better adversarial robustness.
\cite{zhangadversarial} explain AT with causal reasoning, as in our work, they discuss alignment between the data and adversarial distributions, yet their work considers conditionals $p_{\net}(y|\bx)$ and not with marginals.
Orthogonal to all aforementioned works, we show that overfitting is actually linked to the model drastically increasing the discrepancy between natural and adversarial energies.
Our work also connects to~\cite{yu2022understanding}, which ascribes overfitting to data with low loss values. Nevertheless, with our formulation, we can actually show that \emph{low} loss values correspond to attacks that bend the energy even \emph{more} than higher values, see~\cref{fig:fig_quiver} (SAT).

\subsection{EBMs and Robust Classifiers}
\noindent 
The relationship between robust and generative models was explored in~\cite{grathwohl2019your}, where the Joint Energy-based Model (JEM) reformulates the traditional softmax classifier into an energy-based model and trains a single network for hybrid discriminative-generative modeling. In~\cite{yang2021jem++}, an extension of JEM was introduced to enhance the stability and speed of the training. Subsequently, \cite{zhu2021towards} established an initial link between AT and energy-based models, illustrating how they manipulate the energy function differently but share a comparable contrastive approach. Generative capabilities of robust classifiers have been studied in other works~\cite{foret2021sam,wang2022aunified,yang2023towards,yang2023mebm} and even employed in inverse problems~\cite{rojas2021inverting} or controlled image synthesis~\cite{rouhsedaghat2022magic}. However, despite this recent discovery, there is still a lack of extensive research exploring this connection in the context of AT. A recent approach addresses single-step AT \cite{10485467}, while \cite{mirza2024shedding} investigates multi-step AT. We further extend this line of work and provide a deeper understanding of overfitting in AT by analyzing the dynamics of the energy landscape.

\subsection{Weighting the Samples in Adversarial Training} 
\noindent MART~\cite{wang2019improving} started a line of research that shows improvement by weighting the samples in AT. GAIRAT~\cite{zhang2020geometry} follows the trend, although it was shown to be non-robust~\cite{hitaj2021evaluating}. There are several other approaches, such as the continuous probabilistic margin (PM)~\cite{liu2021probabilistic} or weighting with entropy~\cite{kim2021entropy}. Lately, \cite{zhang2023memorization} proposed an approach that applies self-supervised learning to alleviate RO, using memory banks to avoid memorization.

\section{Reconnecting Attacks with the Energy}\label{sec:method}
\noindent In this section, we first define energy as derived from discriminative models. We then outline adversarial attack settings in a white-box scenario before exploring data density modeling and standard discriminative classifiers using EBMs.

\minisection{Discriminative Models as EBM}  
EBMs \cite{lecun2006tutorial} are based on the assumption that any probability density function $p(\bx)$ can be defined through a Boltzmann distribution as $p_{\net}(\bx) = \frac{\exp{(-\Ex)}}{Z(\net)}$, where $\Ex$ is the energy function, mapping each input $\bx$ to a scalar, and $Z(\net) = \int_{}^{}\exp(-\Ex) \,d\bx$ is the normalizing constant s.t. $p_{\net}(\bx)$ is a probability density function.
The joint distribution \( p_{\net}(\bx, y) \) can also be modeled using an energy-based formulation. This allows us to define the corresponding discriminative classifier through the energy function and normalization constant as:
\begin{equation}
\begin{aligned}
\label{eq:join2}
p_{\net}(y|\bx) = \frac{p_{\net}(\bx,y)}{p_{\net}(\bx)} = \frac{\exp{(-\Exy)} Z_{\net}}{ \exp{(-\Ex)} \hat{Z}_{\net}} \\= \frac{\exp{(\net(\bx)[y])}}{\sum_{k=1}^{K} \exp{\net}(\bx)[k]},
\end{aligned}
\end{equation}
where $\hat{Z}_{\net}$ is the normalizing constant of $p_{\net}(\bx,y)$, $Z_{\net} = \hat{Z_{\net}}$~\cite{zhu2021towards} and $\net[i]$ is $i^{th}$ logit.
Observing \cref{eq:join2}, we can deduce the definition of the energy functions as:
\begin{equation}
\begin{aligned}
\Exy = - \log{\exp{ (\net(\bx)[y]) } }\; \text{ and } \\\Ex = - \log{ \sum_{k=1}^{K} \exp{ (\net(\bx)[k]) } } \; .
\label{eq:energy}
\end{aligned}
\end{equation}
This framework offers a versatile approach to consider a generative model within any DNN by leveraging their logits~\cite{grathwohl2019your}.

\noindent Following~\cite{zhu2021towards} and \cref{eq:energy}, we get the cross-entropy (CE) loss as $\Loss_{\text{CE}}(\bx,y;\net)=-\log\big(p_{\net}(y | \bx)\big)=-\net(\bx)[y]+ \log{ \sum_{k=1}^{K} \exp{ (\net(\bx)[k]) } }$ and thus in terms of energy:
\begin{equation}
\begin{aligned}
\Loss_{\text{CE}}(\bx,y;\net)= \underbrace{-\net(\bx)[y]}_{\Exy}+ \underbrace{\log{ \sum_{k=1}^{K} \exp{ (\net(\bx)[k]) } }}_{-\Ex}\\
=\Exy - \Ex.
\label{eq:ce-energy}
\end{aligned}
\end{equation}

\begin{figure}[t]
    \centering
        \centering
        \begin{overpic}[width=0.94\linewidth]{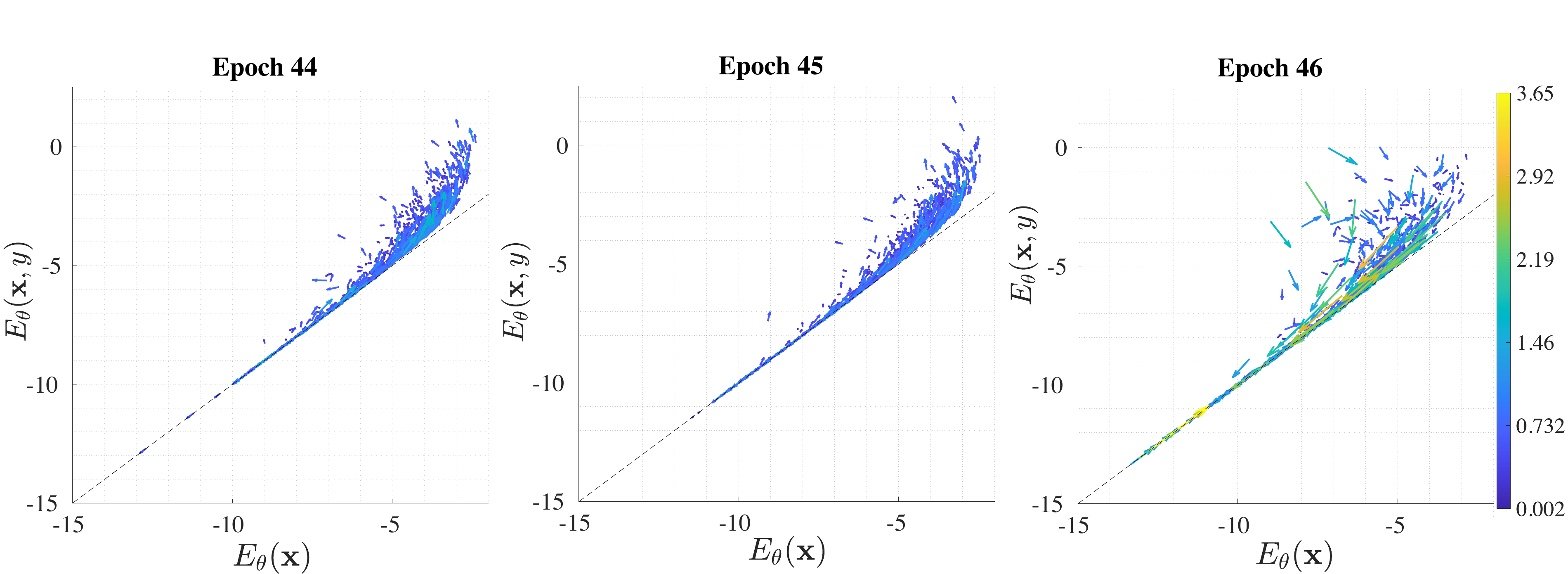}
            \put(5,36.5){\color{gray}\footnotesize{-------------------- RS-FGSM~\cite{wong2020f} - Single-step training -----------------}}
            \put(84.5,31.5){\color{black}\tiny{(CO occured)}}
        \end{overpic}
        \vspace{0.15cm}

        \begin{overpic}[width=0.95\linewidth]{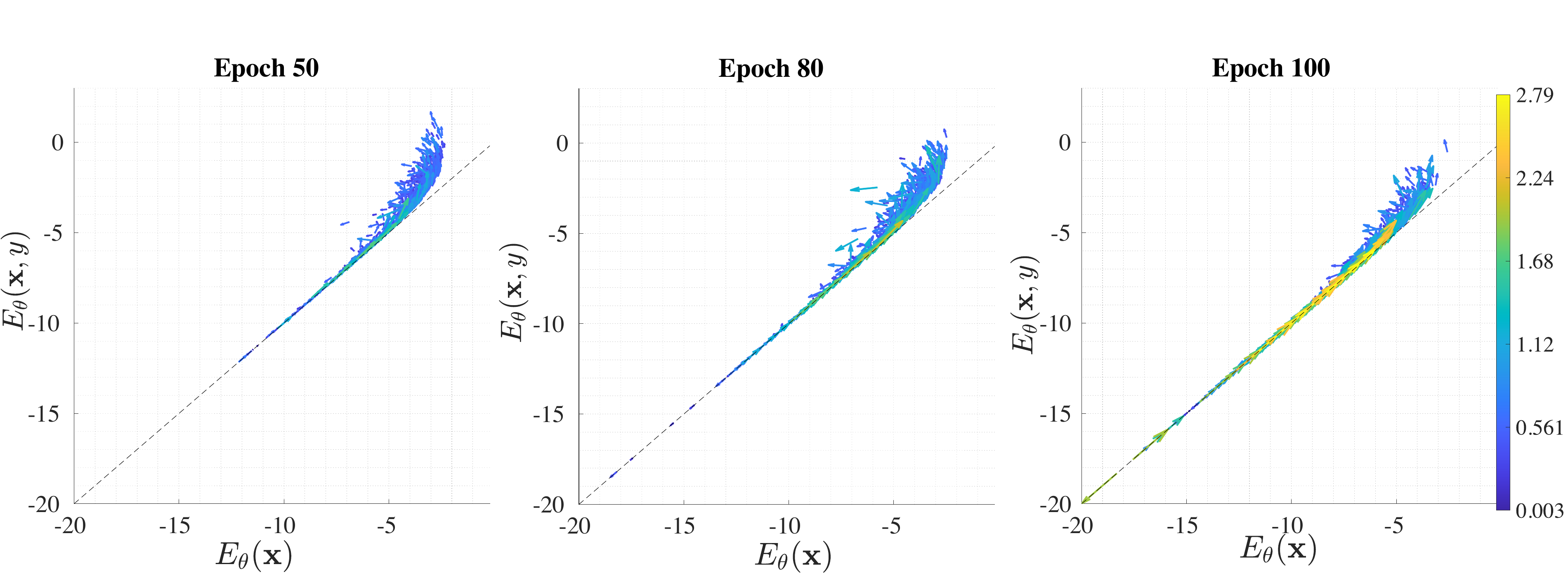}
             \put(5,35){\color{gray}\footnotesize{-------------------- SAT~\cite{madry2017towards} - Multi-step training ---------------------------}}
            \put(84.8,32){\color{black}\tiny{\big(Overfitted(RO)\big)}}
        \end{overpic}
        \vspace{0.15cm}

        \begin{overpic}[width=0.95\linewidth]{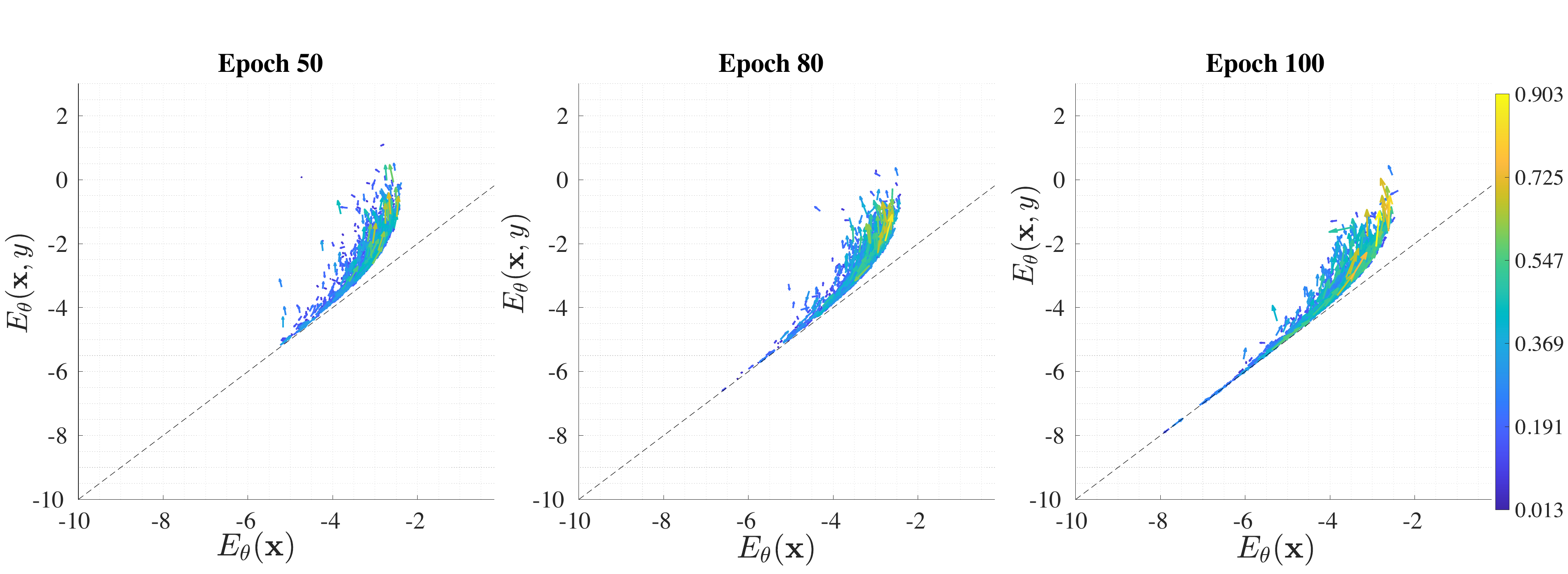}
          \put(5,35){\color{gray}\footnotesize{-------------------- TRADES~\cite{zhang2019theoretically} - Multi-step training -----------------------}}
        \end{overpic}
\caption{Quiver plot illustrates the energy shift between the original and its corresponding adversarial sample. Each arrow, scaled for better visualization, originates from the base—representing $[\Ex, \Exy]$—and points to the tip—indicating $[E_{\net}(\bxa), E_{\net}(\bxa, y)]$. The arrow color encodes the Euclidean distance between these points, reflecting the intensity of energy change, also written as $\norm{  [\Delta \Ex, \Delta \Exy] }_2$. Results are shown for a subset of CIFAR-10 training data at different stages --- note the axes across figures have different ranges for clarity. The dashed black line represents zero cross-entropy, where $\Ex = \Exy$. RS-FGSM is shown just before and after CO, while TRADES and SAT are shown at epochs 50, 80, and 100.}
    \label{fig:fig_quiver}
\end{figure}

\noindent By definition \cref{eq:ce-energy} $\geq 0$ and the loss is zero when $\Exy= \Ex$. To see how the loss used in adversarial attacks induces different changes in energies, we can consider the maximization of~\cref{eq:ce-energy} performed during \emph{untargeted} PGD.

When analyzing multi-step approaches such as PGD, we find that it shifts the input by two terms $\nabla_{\bxa} E_{\net}(\bxa,y)-\nabla_{\bxa}E_{\net}(\bxa)$: a \emph{positive} direction of $\Exy$ and a \emph{negative} direction $\Ex$. 
As found by~\cite{beadini2023exploring}, untargeted PGD finds input points that fool the classifier yet are even more likely than natural data from the perspective of the  classifer, leading to high joint energy but low marginal energy samples. 
To make a connection with denoising score-matching~\cite{song2019generative} and diffusion models~\cite{dhariwal2021diffusion}, we can see how PGD is heavily biased by the score function, i.e. $\nabla_{\bx} \log p_{\net}(\bx )$ since $ \nabla_{\bx} \log p_{\net}(\bx ) = \nabla_{\bx}-\Ex  - \nabla_\bx \log Z_{\net} = -\nabla_\bx \Ex$ where the last identity follows since $\nabla_{\bx}\log Z_{\net}=0$.
On the contrary, it is interesting to reflect on how the dynamic is flipped for \emph{targeted} attacks: assuming we target $y_t$, $-\nabla_{\bxa} E_{\net}(\bxa,y_t)+\nabla_{\bxa}E_{\net}(\bxa)$, the optimization lowers the joint energy yet produces new points in the opposite direction of the score---out of distribution.
To empirically validate this, we evaluate non-robust models and present in~\cref{fig:histogram-energy} the marginal and joint energy distributions under various targeted and untargeted attacks. Multi-step untargeted attacks, such as PGD~\cite{madry2017towards}, PGD-KL~\cite{zhang2019theoretically}, and APGD~\cite{croce2020reliable}— which maximize CE loss or KL divergence—consistently shift the marginal energy $\Ex$ leftward, reflecting decreased overall confidence, while pushing $\Exy$ to the right as they reduce $p(y|\bx)$—consistent with their goal of making the correct label less likely. We can see a similar pattern in Diff-PGD~\cite{xue2023diffusion}, which uses Diffusion Models trained with coupled clean and adversarial examples with PGD to generate adversarial PGD-like samples. Interestingly, for untargeted FGSM~\cite{goodfellow2014explaining}, which also maximizes the CE loss, the $\Ex$ is shifted to the right. This might be due to its one-step nature, which is a coarse approximation of the loss maximization. Targeted attacks like APGD-T~\cite{croce2020reliable} moves the $\Ex$ energy to the right to push $\Exy$ to the target label, thereby creating points that are more out-of-distribution compared to natural samples which is a behavior already noted in~\cite{wang2022aunified}. Margin-based attacks such as CW~\cite{carlini2017towards} and APGD-DLR~\cite{croce2020reliable} manipulate the logits with minimal distortion, the marginal energy $\Ex$ remains nearly unchanged or shifts only slightly rightward. Similarly, the black-box Square~\cite{andriushchenko2020square} conducts a localized random search to cross the decision boundary with minimal disruption to the logit distribution.

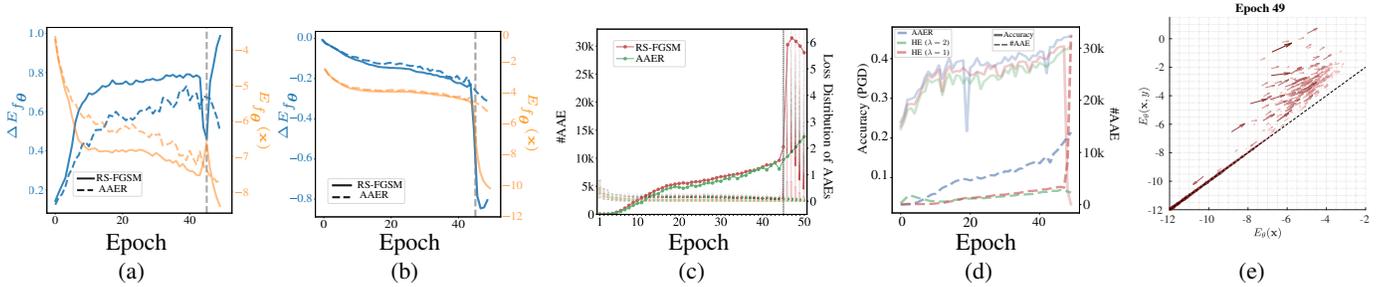
\begin{figure*}[t]
    \centering
     \subfloat[]{
    \begin{overpic}[width=.18\linewidth]{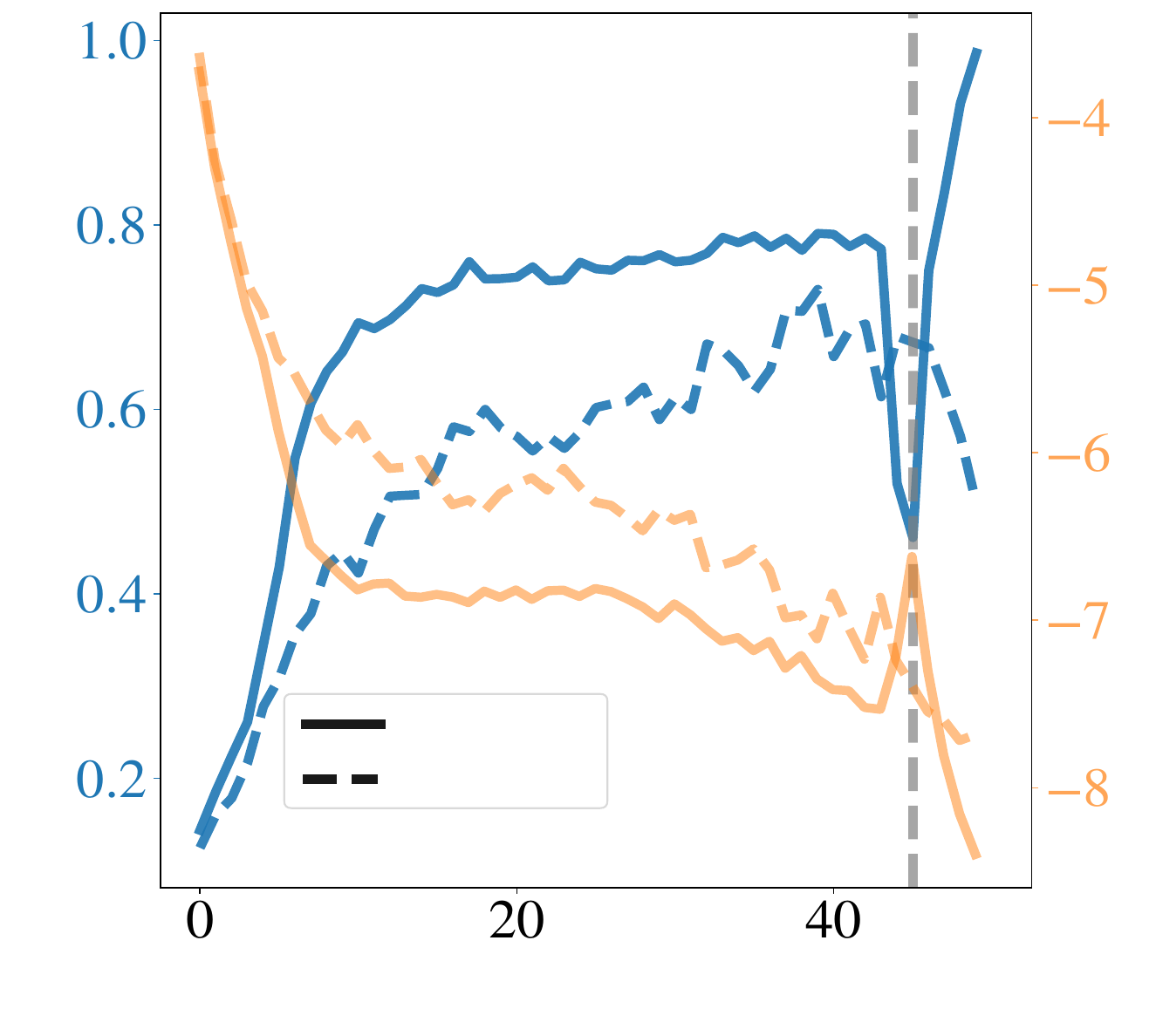}
        \put(34.1,25.8){\scalebox{.4}{RS-FGSM}}
        \put(35,21.5){\scalebox{.4}{AAER}}
        \put(-2, 43){\rotatebox{90}{\color{plotBlue}\tiny{$\Delta E_{\net}$}}}
        \put(97,63){\rotatebox{-90}{\color{plotOrange}\tiny{$\Ex$}}}
        \put(38,-2){\small{Epoch}}
    \end{overpic}
    \label{fig:aae_b}
    }
    \subfloat[]{
    \begin{overpic}[width=.18\linewidth]{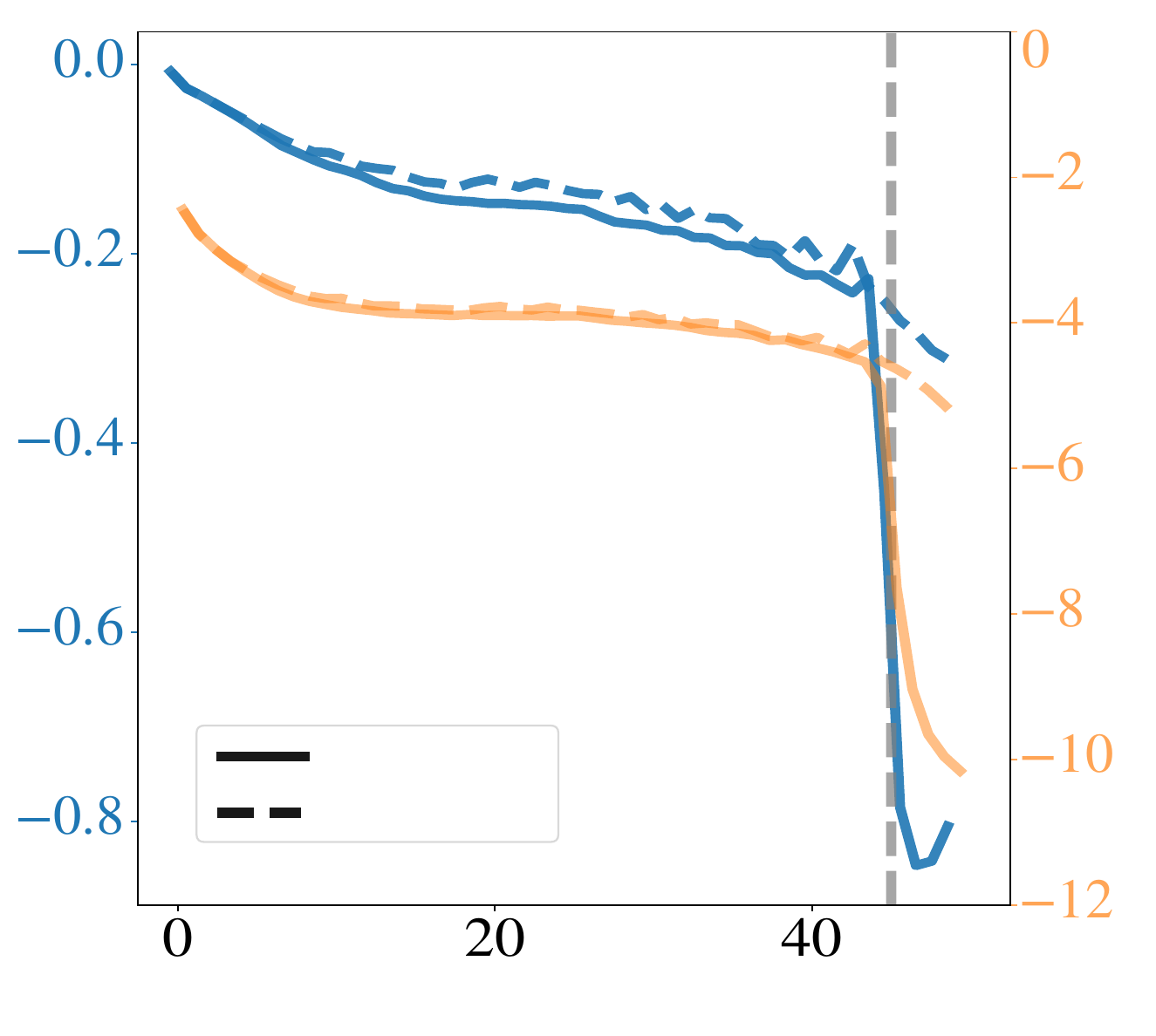}
        \put(29.1,23.2){\scalebox{.4}{RS-FGSM}}
        \put(30,18.8){\scalebox{.4}{AAER}}
         \put(-3, 43){\rotatebox{90}{\color{plotBlue}\tiny{$\Delta E_{\net}$}}}
        \put(95,62){\rotatebox{-90}{\color{plotOrange}\tiny{$\Ex$}}}
        \put(38,-2){\small{Epoch}}
    \end{overpic}
    \label{fig:aae_c}
    }
     \subfloat[]{
    \begin{overpic}[width=.20\linewidth]{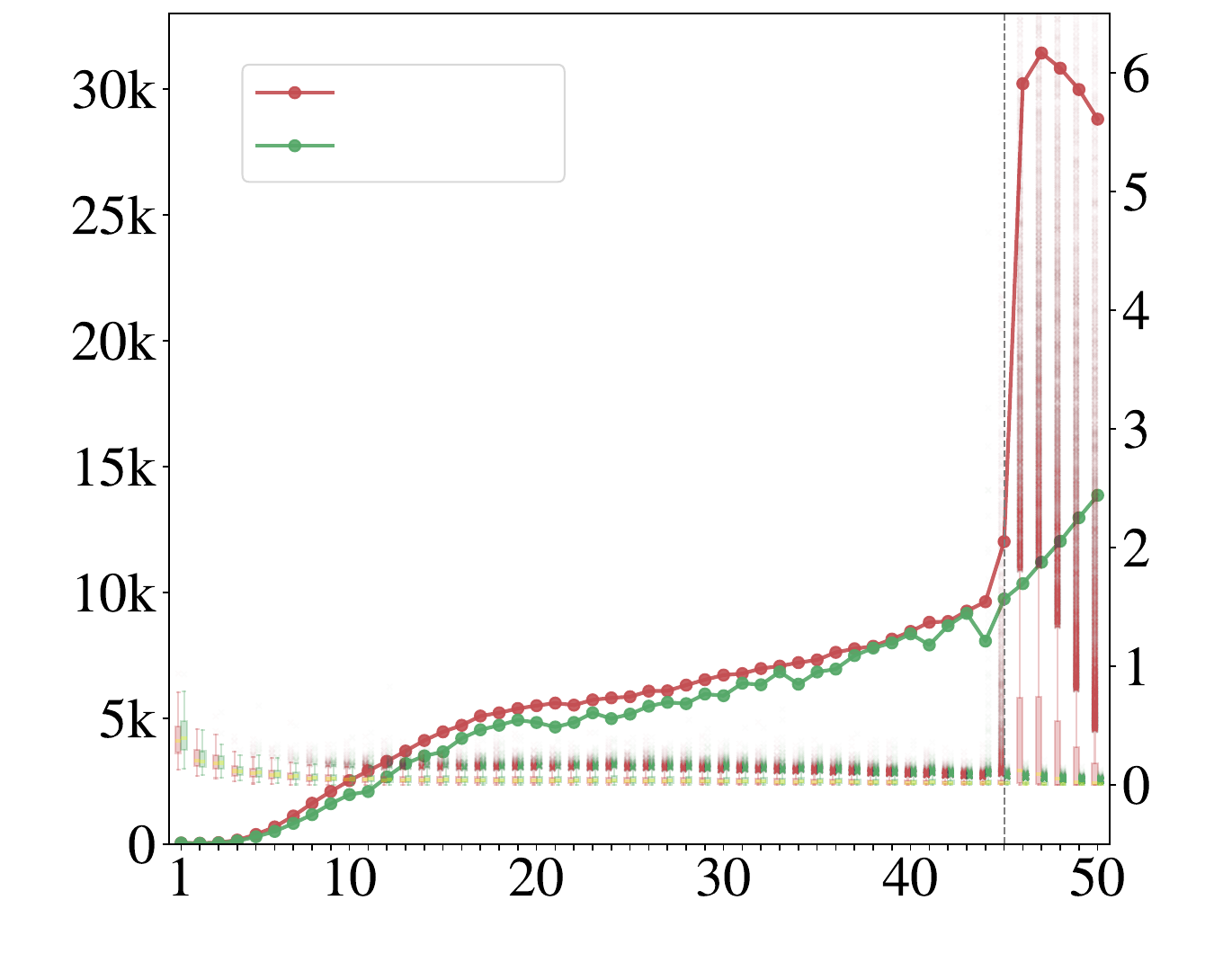}
        \put(28,71){\scalebox{.4}{RS-FGSM}}
        \put(28,67){\scalebox{.4}{AAER}}
        \put(-1, 37){\rotatebox{90}{\tiny{\#AAE}}} 
        \put(97,70){\rotatebox{-90}{\tiny{Loss Distribution of AAEs}}} 
        \put(40,-2){\small{Epoch}}
    \end{overpic}
    \label{fig:aae_a}
    }
    \subfloat[]{
    \begin{overpic}[width=.18\linewidth]{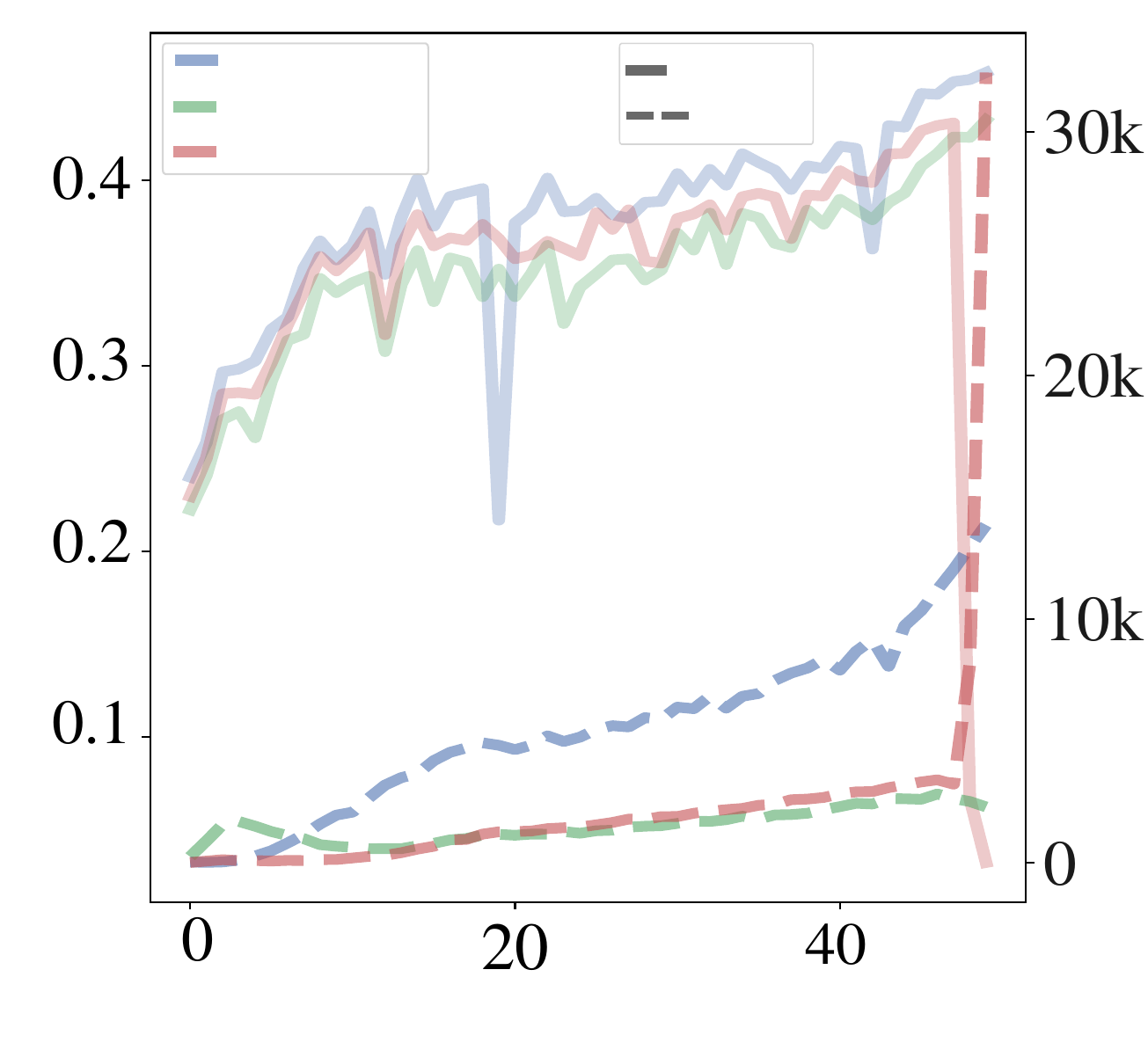}
        \put(21,85.5){\scalebox{.3}{AAER}}
        \put(21,81.5){\scalebox{.3}{HE ($\lambda=2$)}}
        \put(21,77.5){\scalebox{.3}{HE ($\lambda=1$)}}
        \put(58.5,84.5){\scalebox{.3}{Accuracy}}
        \put(61,81){\scalebox{.3}{\#AAE}}
        \put(-1, 37){\rotatebox{90}{\tiny{Accuracy (PGD)}}} 
        \put(102,56){\rotatebox{-90}{\tiny{\#AAE}}}
        \put(39,-2){\small{Epoch}}
    \end{overpic}
    \label{fig:aae_d}
    }
    \subfloat[]{
    \begin{overpic}[width=.18\linewidth]{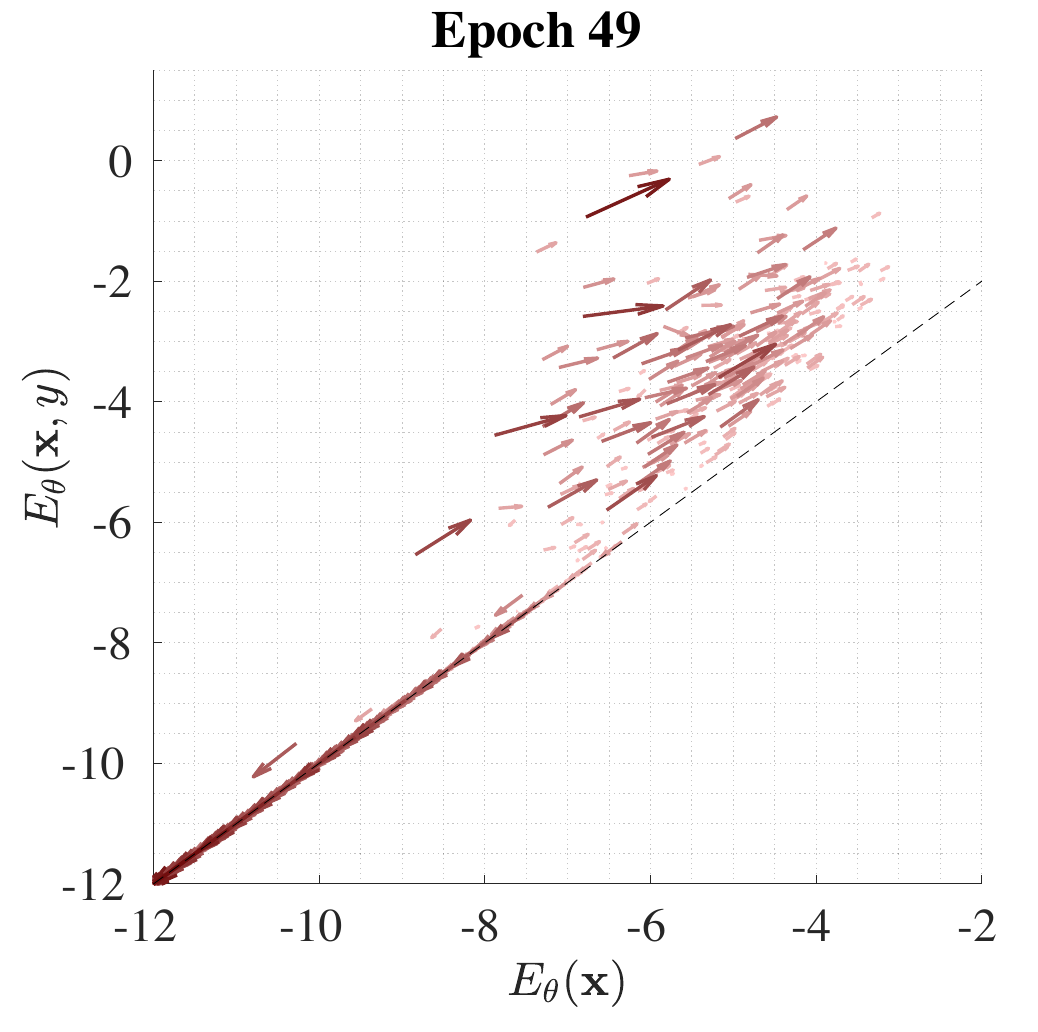}
    \end{overpic}
    \label{fig:aae_e}
    }
    \caption{\textbf{(a, b)} Mean $\Delta \Ex$ values (left axis) and mean energy values for original samples $\Ex$ (right axis) across training epochs for AAEs and NAEs, respectively. \textbf{(c)} The number of AAEs (left axis) and the distribution of their losses (right axis) are tracked over the course of training. The dashed line indicates the onset of CO and we compare RS-FGSM (which suffers from CO) and AAER.   
    \textbf{(d)} Robust accuracy (PGD) on the test set (left axis) and number of AAEs (right axis) over training.  
    \textbf{(e)} Quiver plot of AAEs generated when training with HE ($\lambda = 0.2$) at later epochs, showing high-energy, high-loss AAEs that do not lead to CO.  
    All analyses are conducted on CIFAR-10.}  
    \label{fig:aae}
\end{figure*}

\section{Understanding Overfitting with Energy}

We find the energy plays a key factor in understanding the behavior of overfitting in AT,  both in the context of RO and CO. To show this, we performed an analysis in which we compared the energies of the samples in \emph{training set} with their corresponding adversarial counterparts in each epoch during AT. Given an input image $\bx$ and its adversarial counterparts $\bxa$, we measure the difference between their marginal energies $\Ex-\Exp$, denoted by $\Delta \Ex$ and the joint energy relative to the ground truth label $y$, $\Exy-\Exyp$, denoted by $\Delta \Exy$. We further quantify the intensity of this shift using the $L_{2}$ norm, $\norm{  [\Delta \Ex, \Delta \Exy] }_2$, which captures the variation in both the marginal and joint energies relative to the ground truth as samples transition from original to adversarial, as shown in \cref{fig:fig_quiver}.

\subsection{Catastrophic Overfitting}
\noindent We begin by analyzing the single-step AT approach using RS-FGSM\cite{wong2020f} on CIFAR-10 and CIFAR-100 datasets. RS-FGSM is one of the earlier single-step AT methods and provides a useful baseline for investigating CO, as this method is known to suffer from  CO for longer training. As seen in \cref{subfig:overfit-CO}, we analyze the trend of mean $\Delta \Ex$ over all training samples and find that the training progress is split into two distinct phases. In the initial phase, the energies of the original and adversarial counterparts are comparable. However, as soon as CO occurs, $\Delta \Ex$ starts to increase significantly, because the energy of the original samples, \(\Ex\), increases significantly compared to the adversarial samples \(\Exp\), i.e. $\Ex \gg \Exp$ for most samples,  as shown in~\cref{fig:fig_quiver}~(RS-FGSM). We further investigated this behavior by analyzing how $\Delta \Ex$ behaves differently for AAEs and NAEs, providing insight into sample-specific dynamics during CO.

\minisection{Abnormal Adversarial Examples} 
\noindent During AT when the inner maximization seeks the worst-case perturbation \(\boldsymbol{\delta} \in \Delta\) that maximizes the loss \(\mathcal{L}(\net(x + \boldsymbol{\delta}), y)\) as shown in \cref{eq:adversarial_training}, the resultant image should have higher loss than the original image. However, earlier studies~\cite{vivek2020single} discovered this is not always true, indeed, they found certain instances that generated adversarial images while training were not adversarial. \cite{lin2024eliminating} formally referred to such examples as abnormal adversarial examples due to this abnormal behavior.

\subsubsection{Energy Perspective on AAEs}{\label{energy_method_ss}}The insights from \cite{lin2024eliminating} provide a deeper understanding of the sample-wise dynamics under the phenomenon of CO. To further investigate the behavior of $\Delta \Ex$, particularly its sharp increase at the onset of CO as observed in ~\cref{subfig:overfit-CO}, we analyze it separately for the AAEs and NAEs, as shown in \cref{fig:aae_b} and \cref{fig:aae_c}, respectively.
Once CO sets in, we observe an abrupt surge in the number of AAEs, with $\Ex > \Exp$ for almost all of them which causes $\Delta \Ex$ to increase. In contrast, approaches that mitigate CO — such as incorporating Abnormal Adversarial Examples Regularization (AAER)~\cite{lin2024eliminating} into RS-FGSM results in — $\Delta \Ex$ starts to decrease towards the end of the training, precisely when a large number of AAEs would otherwise be generated.

\subsubsection{How Many AAEs Trigger CO}
\begin{figure}[t]
    \centering
        \begin{overpic}[width=0.95\linewidth]{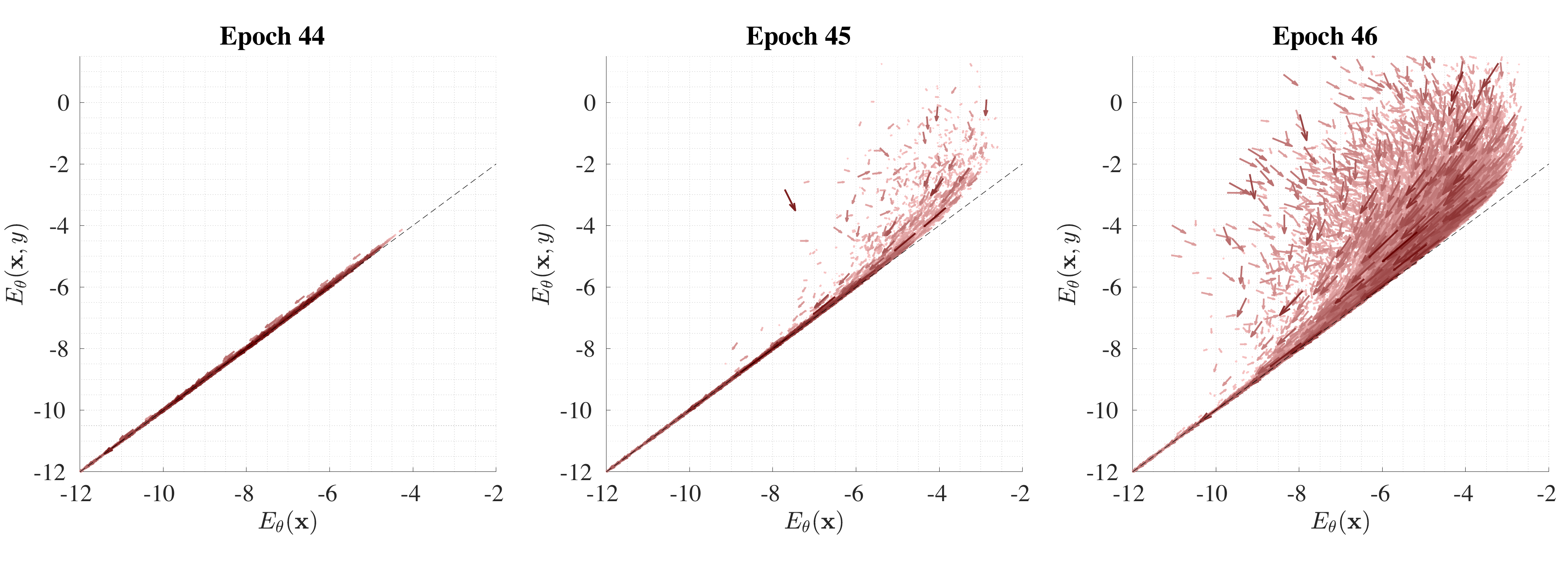}
            \put(5,36.5){\color{gray}\footnotesize{-------------------- RS-FGSM~\cite{wong2020f} - Single-step training -----------------}}
            \put(88,33.5){\color{black}\tiny{(CO occured)}}
        \end{overpic}
        \vspace{0.15cm}
        
        \begin{overpic}[width=0.95\linewidth]{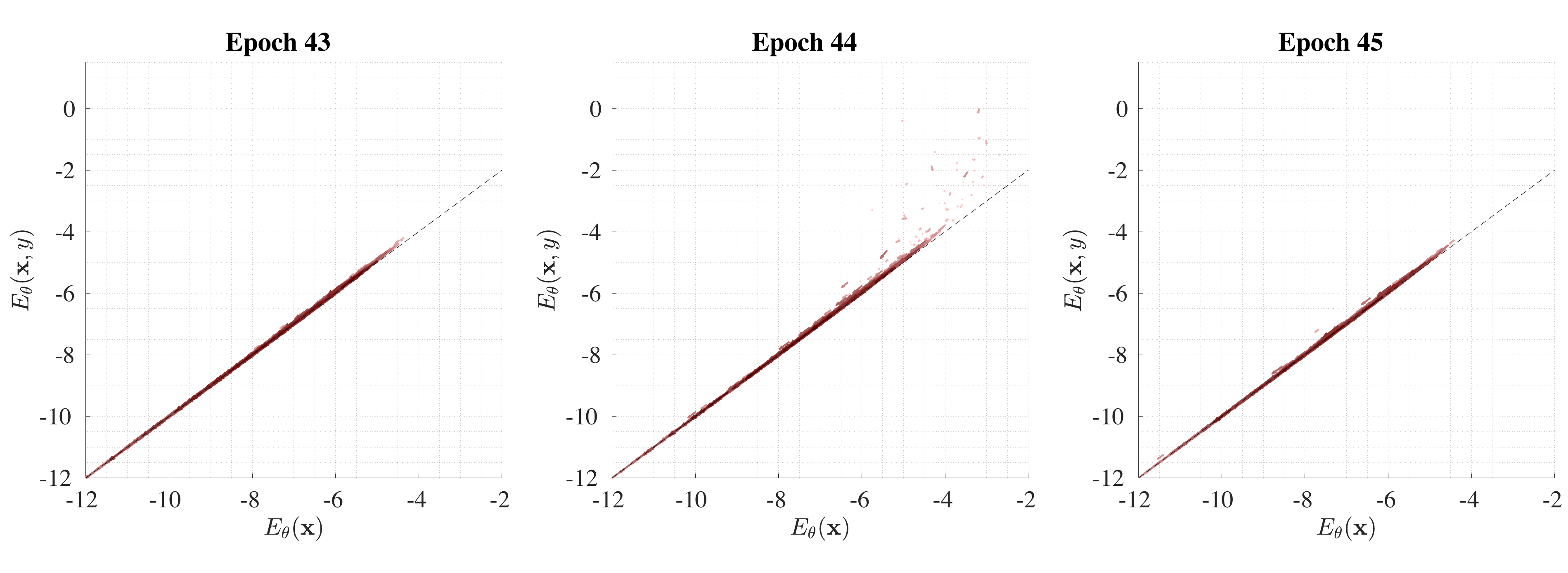}
            \put(5,35.5){\color{gray}\footnotesize{----------------------AAER~\cite{lin2024eliminating} - Single-step training --------------------}}
        \end{overpic}
        \vspace{0.15cm}
        
        \begin{overpic}[width=0.95\linewidth]{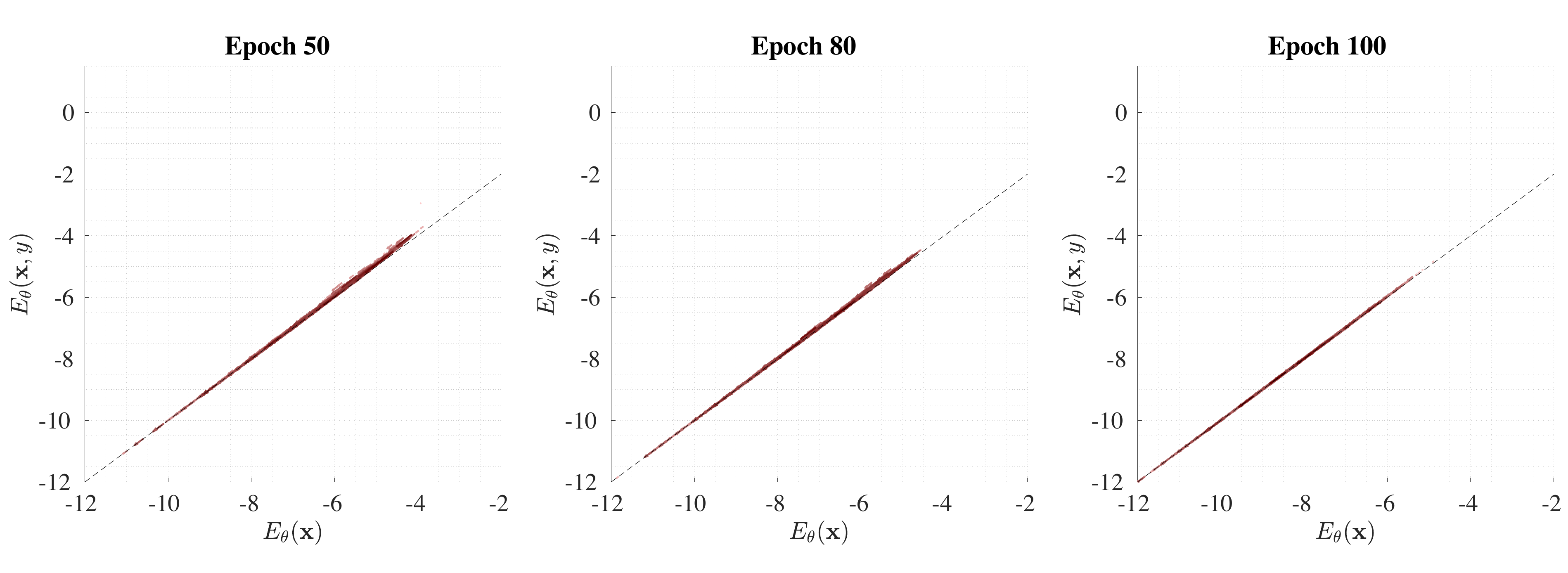}
            \put(5,35.2){\color{gray}\footnotesize{---------------------- SAT~\cite{madry2017towards} - Multi-step training -------------------------}}
        \end{overpic}
    
    \caption{Quiver plot now only showing all the AAEs present in the training set. For RS-FGSM, just before CO, AAEs appear at high energy levels with increased loss, deviating from the dashed line. In contrast, for methods that do not undergo CO, AAEs consistently exhibit low energy and loss.}
    \label{fig:abnormal_quiver}
\end{figure}

In \cref{fig:aae_a}, we show the number of AAEs generated at each epoch during training. We compare AAER with RS-FGSM, where RS-FGSM suffers from CO at later epochs. 
Although both methods exhibit a steady increase in AAEs over time, with nearly identical overall counts, a slight increase in AAEs is observed just before CO occurs. This observation prompts an important question: \textit{How the AAEs generated at CO differ from those produced earlier in training?}\\  
To answer this, we plot the sample-wise energies of AAE generated while training with RS-FGSM as shown in \cref{fig:abnormal_quiver}~(RS-FGSM) and observe that the AAEs generated earlier in training, before CO, exhibit low energy values and correspondingly have lower loss, as shown in \cref{fig:aae_a}. However, just before the CO phase, the AAEs shift to higher energy values and their loss also increases significantly. This observation can be confirmed from \cref{fig:aae_b}  where we show that right before CO, the mean energy across all abnormal samples increases, and \cref{fig:aae_b} where we observe certain AAE samples appearing with higher loss\footnote{Here the loss is calculated on the corresponding original samples of AAEs.}. In particular, when training with AAER, shown in \cref{fig:abnormal_quiver}~(AAER), while some high-energy AAEs briefly appear at the end of training, the method successfully suppresses them, preventing recurrence in subsequent epochs. Furthermore, our analysis of multi-step approaches, such as SAT~\cite{madry2017towards}, reveals that in contrast to earlier beliefs that CO does not occur in multi-step approaches due to the lower number of AAEs, all AAEs throughout training consistently exhibit low energy as shown in \cref{fig:abnormal_quiver}~(SAT). We emphasize that the number of AAEs is not the primary factor in triggering CO; it is their energy or their corresponding loss that plays a crucial role.  
\subsubsection{AAEs when the Inner Maximization is not CE Loss}Since we observe that $\Ex < \Exp$ for most NAEs, a straightforward approach for mitigating CO is to incorporate the High-Energy Regularizer (HE)~\cite{beadini2023exploring} to generate adversarial samples with higher energy during the inner maximization of Eq.~\eqref{eq:adversarial_training}. Applying this regularizer to RS-FGSM, we find that with an appropriate choice of $\lambda$\footnote{$\lambda$ controls the influence of the regularizer on the overall loss.}, CO can be effectively prevented, as shown in \cref{fig:aae_d}. Notice how with lower $\lambda$, the model suffers from CO, despite generating significantly fewer AAEs before CO occurs — further supporting the observation that the number of AAEs alone does not trigger CO.
Interestingly, when analyzing the AAEs produced using the approach with higher $\lambda$, \cref{fig:aae_e}, we find that although these AAEs exhibit high energy and also higher loss, CO does not occur. This leads us to an important question: \textit{How should AAEs be defined, if the inner maximization objective is not CE loss}?\\ For instance, the inner maximization is KL divergence for TRADES~\cite{zhang2019theoretically} or the Guided Adversarial Margin Attack (GAMA) for Guided Adversarial Training (GAT)~\cite{sri2020guided}. Notably, these methods typically combine the CE loss with an additional regularizer in the outer minimization while training. In both the case of TRADES and GAT, we observe a higher number of AAEs, with a significant portion exhibiting both high energy and high loss, yet they do not suffer from CO. 
However, when we generate adversarial samples using the CE loss for these models, we observe a familiar pattern: the AAEs generated exhibit lower energy and loss, similar to our earlier observations. This emphasizes that AAEs are not directly the cause of CO as previously noted by~\cite{lin2024eliminating}; instead, they serve as a diagnostic indicator reflecting the underlying state of the classifier. In other words, the distortion in the classifier’s decision boundaries is the true driver of CO and the AAEs are the byproduct of this distortion.\\

\subsection{Robust Overfitting}

\noindent When using a multi-step AT approach like SAT~\cite{madry2017towards}, we find a totally different trend, although the training is still divided into two phases where the first one is similar to the single-step approach one, suggesting that original and adversarial samples energies still exhibit comparable values. 
In contrast, during the second phase, $\Ex$ and $\Exp$ trends begin to diverge from each other, simultaneously observing an increase in test error, as shown in \cref{subfig:overfit-CO},  suggesting RO is occurring.
In this case, $\Delta \Ex$ shows a steep decrease, behaving completely different w.r.t. CO.
Thus, to alleviate RO, it seems imperative to maintain similarity in energies between original and adversarial samples, thereby smoothing the energy landscape around each sample. Interestingly, reinterpreting TRADES~\cite{zhang2019theoretically} as EBM reveals that TRADES is essentially achieving the desired objective, towards a notable mitigation of RO.

\begin{figure*}[t]
\centering
\subfloat[]{%
\begin{overpic}[height=1.2in]{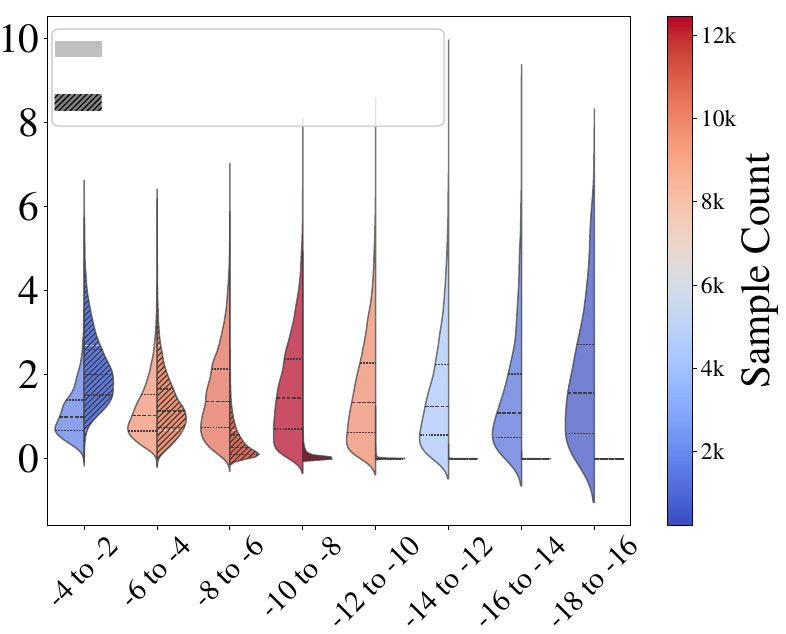}
    \put(36,-7){$\Ex$}
    \put(14.5,66){\scalebox{.4}{Cross-Entropy Loss}}
    \put(14.5,73){\scalebox{.4}{$\norm{  [\Delta \Ex, \Delta \Exy] }_2$}}
\end{overpic}
\vspace{1.5mm}
\label{fig:binsDeltaEx}}
\subfloat[]{%
\begin{overpic}[height=1.2in]{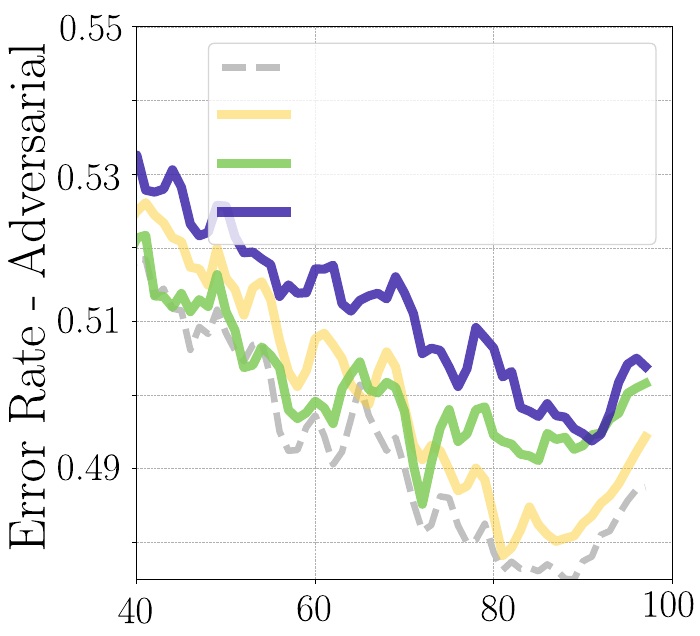}
    \put(47,-5.5){\small{Epoch}}
    \put(41.6,80){\scalebox{.4}{ SAT (with all data)}}
    \put(41.6,74) {\scalebox{.4}{ w/o Low Energy}}
    \put(42.3,66.3) {\scalebox{.4}{w/o Incorrect (MART)}}
    \put(42.3,59.3) {\scalebox{.4}{w/o High $\Ex$ \& Correct}}
    
\end{overpic}
\label{fig:ablation-HE-samples}}
\subfloat[]{%
\begin{overpic}[height=1.2in]{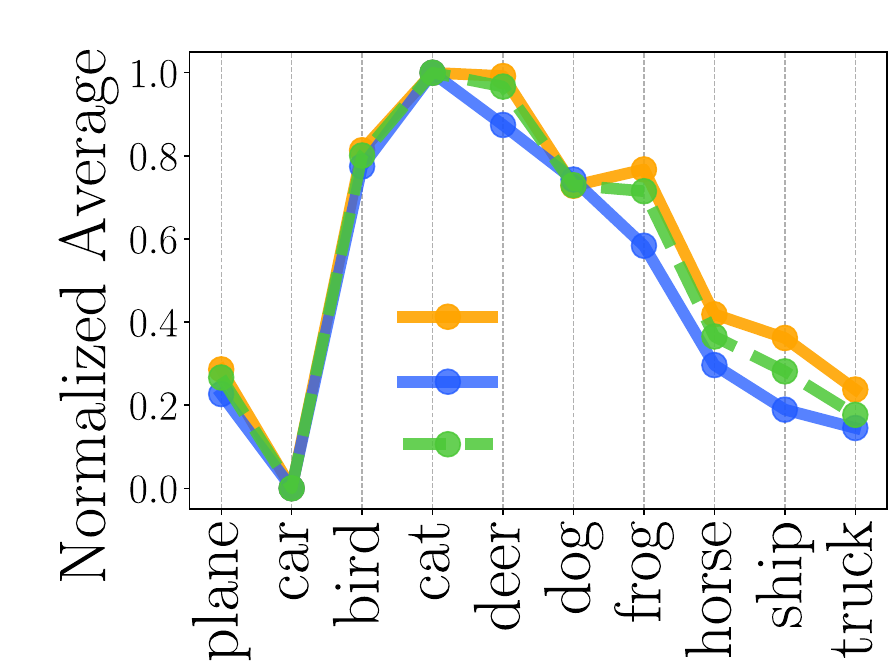}
    \put(42,-5){\small{Classes}}
    \put(57,38){\scalebox{.6}{$\Ex$ }}
    \put(57,31) {\scalebox{.6}{Error}}
    \put(57,24) {\scalebox{.6}{Entropy}}
\end{overpic}
\vspace{1.5mm}
\label{fig:ent-ex}}
\subfloat[]{%
\begin{overpic}[height=1.1in]{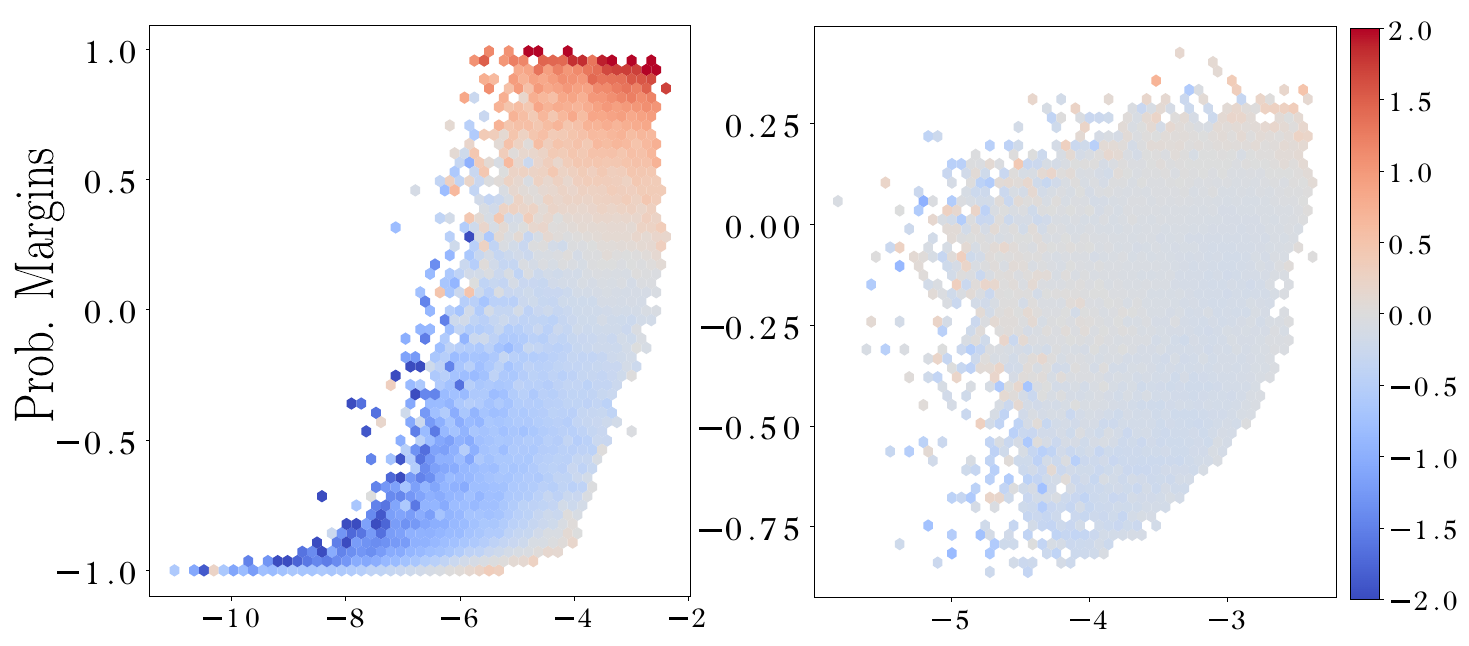}
    \put(20,-4){$\Ex$}
    \put(66,-4){$\Ex$}
    \put(10,45){\scalebox{.7}{Overfitted Model (RO)}}
    \put(58,45){\scalebox{.7}{MAIL-AT (No RO)}}
    \put(100,34){\rotatebox{270}{{$\Delta\Ex$}}}
\end{overpic}
\label{fig:pm-ex}}
\caption{\textbf{(a)} Distribution of Cross-Entropy loss and  $\norm{  [\Delta \Ex, \Delta \Exy] }_2$ across bins of $\Ex$ at the end of the training calculated on an overfitted model (RO). The colorbar shows the number of samples in each bin. \textbf{(b)} Not perturbing high-energy samples (correctly classified) increases robust error akin to not perturbing incorrectly classified samples, shown in~\cite{wang2019improving}. \textbf{(c)} The relationship between error rate, entropy, and energy~\cite{losch2023onallexamples} indicates that samples with higher energy tend to transfer robustness more effectively across classes. \textbf{(d)} Hexbin plots showing the relationship between Probabilistic Margins (PMs) (y-axis) and energy \( \Ex \) (x-axis), where each hexbin is colored by the average change in energy \( \Delta \Ex \) of the samples within it. Results are shown for training samples from the final epoch of models trained with SAT (exhibiting RO) and MAIL-AT (no RO); notice that the latter exhibits \( \Delta \Ex \) values closer to zero, higher energy samples and PMs without extreme values.}
\label{fig:energy-plots}
\end{figure*}
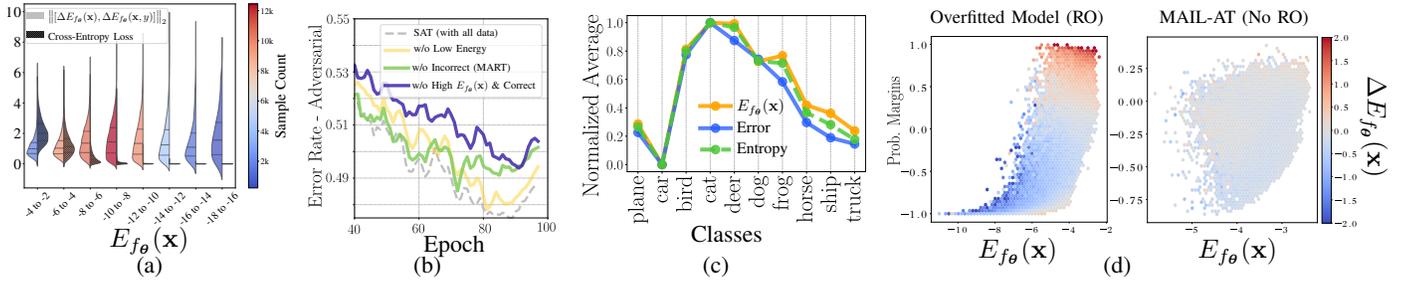

\subsubsection{Interpreting TRADES as Energy-based Model} Going beyond prior works\cite{grathwohl2019your,zhu2021towards,wang2022aunified,beadini2023exploring}, we reinterpret the TRADES objective~\cite{zhang2019theoretically} as an EBM. TRADES loss is as follows:
\begin{equation}
    \min_{\net} \biggl[ \Loss_{\text{CE}}\bigl(\net(\bx),y\bigr)+\beta \max_{\pert \in \Delta} \on{KL}\Big(
    p(y|\bx)\Big|\Big|
    p(y|\bxa)\Big)\biggr],
    \label{eq:trades}
\end{equation}
where $\Loss$ is the CE loss, $p(y|\bx)$ is from~\cref{eq:join2} and $\on{KL}(\cdot,\cdot)$ is the KL divergence between the conditional probability over classes $p(y|\bx)$ that acts as reference distribution and the probability over classes for generated points $p(y|\bxa)$.
\vspace{5pt}
\begin{proposition} 
KL divergence between two discrete distributions $p(y|\bx)$ and $p(y|\bxa)$ can be interpreted using EBM as:
\begin{equation}
\underbrace{\mathbb{E}_{k\sim p(y|\bx) }\Big[\Exkp -\Exk \Big]}_{\text{\emph{conditional term weighted by classifier prob.}}} + \underbrace{\Ex-\Exp}_{\text{\emph{marginal term}}}.
\label{eq:kl-ebm}
\end{equation}
\label{prop:1}
\end{proposition}

By writing the KL divergence as \cref{eq:kl-ebm}, we can better see the analogies and differences with SAT. Similarly to SAT, TRADES has to push down $\Exp$, yet it does so considering a reference fixed energy value that is given by the corresponding natural data $\Ex$. At the same time, they both have to push up $\Exkp$ yet TRADES attack only increases the loss when $\Exkp>\Exk$ for $k$ classes. Furthermore, a big difference lies in the training dynamics: while AT is agnostic to the dynamics, TRADES uses the classifier prediction as a weighted average: at the beginning of the training $p(y|\bx)$ is uniform, being the conditional part averaged across all classes, so the attack does not really affect any class in particular. Instead, at the end of the training, when $p(y|\bx)$ may distribute more like a one-hot encoding, TRADES will consider the most likely class. 

\subsubsection{AT in function of High vs Low Energy Samples} Several studies have highlighted the unequal impact of samples in AT in both single-step training and multi-step training:~\cite{wang2019improving,zhang2020attacks,Ding2020MMA,lin2024eliminating,huang2023fast}. Looking at multi-step approaches,~\cite{wang2019improving} focuses on the importance of samples in relation to their correct or incorrect classification, while~\cite{liu2021probabilistic,zhang2020geometry} suggest that samples near the decision boundaries are regarded as more critical. We can comprehensively explain such findings, as well as others~\cite{yu2022understanding,losch2023onallexamples} using our framework.
We begin by investigating MART, which employs Misclassification-Aware Regularization (MAR), focusing on the significance of samples categorized by their correct or incorrect classification. We do a proof-of-concept experiment closely resembling MART's~\cite{wang2019improving} where we initially start from a robust model trained with SAT~\cite{madry2017towards}. Unlike~\cite{wang2019improving}, we opted to make subsets based on their energy values. We selected two subsets from the natural training dataset: one comprising high-energy examples but excluding misclassifications; another with low-energy samples of correctly classified examples. All subsets are created considering the initial values from the robust SAT classifier.
We trained again the same networks from scratch without these subsets. Subsequently, we assessed the robustness against PGD~\cite{madry2017towards} on the test dataset. Our findings indicate that removing high-energy correct samples has a similar impact to removing incorrectly classified samples, as shown in~\cref{fig:ablation-HE-samples}. Additionally, we observed that most incorrectly classified samples exhibit higher energies, suggesting that robustness reduction is likely due to their high energy values and not to their incorrect classification. Also for lower energy samples, their adversarial counterparts are not really adversarial, having much lower loss as shown in ~\cref{fig:binsDeltaEx}.
Using this formulation, we can also explain recent research~\cite{losch2023onallexamples} revealing that robustness can be transferred to other classes that have never been attacked during AT. Findings from~\cite{losch2023onallexamples} indicate that the classes that are harder to classify show better transfer of robustness to other classes. Moreover, they found that classes with high error rates happen to have high entropy. Our analysis shows that the same classes with high error rates\footnote{We report probabilistic error rate $1-p(y|\bx)$, unlike hard error rates in~\cite{losch2023onallexamples}.} also display higher energy, as shown in~\cref{fig:ent-ex}. Thus, we can infer that classes with higher energy levels better facilitate robustness transferability.
\subsubsection{$\Delta \Ex$ drops at the onset of RO}{\label{sec:energy_method_ms}} While training with SAT, we see $\Delta \Ex$ dropping as soon model starts to overfit, see~\cref{subfig:overfit-RO}. We analyze $\Delta \Ex$, $\Ex$, and Probabilistic Margins (PMs)\cite{liu2021probabilistic} for all samples for an overfitted model as shown in~\cref{fig:pm-ex}. Interestingly, we observe that samples with negative PMs and low energy are those for which $\Delta \Ex$ becomes negative. When the weighting scheme based on PMs is applied to SAT, as in MAIL-AT~\cite{liu2021probabilistic}, the model no longer overfits, samples no longer have extreme PM values, and $\Delta \Ex$ remains near zero—an effect that appears to emerge as a byproduct of their weighting strategy, despite not being explicitly targeted. \emph{This raises the question of whether the improvement in robustness comes from the absence of samples with extreme PM values or from the change in $\Delta \Ex$?} We find that with our newly proposed approach, the $\Delta \Ex$ is brought close to zero, while the PMs of correctly classified samples remain largely unchanged. This suggests that the improvement in robustness is more closely tied to smoothening the energy surface rather than eliminating samples with extreme PM values.
We also note that MAIL-AT assigns near-zero weights to samples with negative PMs and this alone can improve robustness against untargeted PGD attacks, further discussed in ~\cref{sec:ablation}. This finding is in line with the work from~\cite{yu2022understanding} that identifies that some small-loss data samples lead to RO.

\subsection{Better Robust Models Have Smoother Energy Landscapes} 
\begin{figure}[t]
  \centering
  \begin{overpic}[width=\columnwidth]{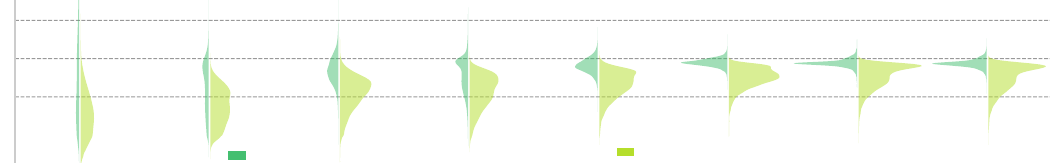}
   \put(4.5,15){\scriptsize{49.25\%}}
   \put(16.5,15){\scriptsize{59.64\%}}
   \put(29.5,15){\scriptsize{53.08\%}}
   \put(41,15){\scriptsize{56.29\%}}
   \put(54,15){\scriptsize{56.17\%}}
   \put(66.5,15){\scriptsize{57.09\%}}
   \put(79,15){\scriptsize{67.73\%}}
   \put(91.5,15){\scriptsize{70.69\%}}
   
   \put(-1.4,13){\scriptsize{+1}}
   \put(-0.8,5.5){\scriptsize{-1}}
   \put(-1.2,9){\scriptsize{ 0}}
   \put(-3.6,4.9){\rotatebox{90}{\scalebox{.7}{$\Delta E_{\net}$}}}
   \put(4,-2.75){\tiny{SAT\cite{madry2017towards}}}
   \put(14,-2.75){\tiny{GAIR-RST\cite{zhang2020geometry}}}
   \put(19.25,-6.5){\scriptsize{+ \faImage}}
   \put(28.5,-2.75){\tiny{TRADES\cite{zhang2019theoretically}}}
   \put(42,-2.75){\tiny{MART\cite{wang2019improving}}}
   \put(45.25,-6.5){\scriptsize{+ \faImage}}
   \put(54,-2.75){\tiny{AWP\cite{wu2020adversarial}}}
   \put(67,-2.75){\tiny{IKL\cite{cui2023decoupled}}}
   \put(78,-2.75){\tiny{IKL\cite{cui2023decoupled}}}
   \put(79.5,-6.5){\scriptsize{+ \faImages}}
   \put(89,-2.75){\tiny{Better DM\cite{wang2023better}}}
   \put(91.5,-6.5){\scriptsize{+ \faImages}}
   \put(25.5,0.8){\scriptsize{$\Ex-\Exp$}}
   \put(60.5,0.8){\scriptsize{$\Exy-\Exyp$}}
  \end{overpic}
  \vspace{10pt}
  \caption{Difference in the energy between natural data $\bx$ and $\bxa$ for \sota methods in adversarial robustness. For each method, we show the signed difference between $\bx$ and $\bxa$ for both $\Ex$ and $\Exy$, on top of each method, we report the robust accuracy from~\cite{croce2020reliable}. The vertical axis is in symmetric \emph{log scale}. The increase in robust accuracy correlates well with $\Delta\Ex$ approaching zero and reducing the spread of the distribution. {\scriptsize + \faImages} indicates training with generated images by~\cite{wang2023better}, while the {\scriptsize + \faImage} indicates training with additional data by~\cite{carmon2019unlabeled} for the CIFAR-10 dataset.
  }
\label{fig:energy_smoothness}
\end{figure}

Smoothness is a well-established concept in robustness, where a smooth loss landscape suggests that for small perturbations $\pert$, the difference in loss $|\Loss_{\net}(\bx) - \Loss_{\net}(\bx+\pert)|$ remains small ($<\epsilon$) w.r.t. the input $\bx$. We show a link between Energy and Loss in~\cref{eq:ce-energy}. PGD-like attacks drastically bend the energy surface---see \cref{fig:histogram-energy}---thereby the model needs to reconcile the adv. energy with the natural energy. This reconciliation yields the smoothness. The intuition is that classifiers may tend towards the data distribution to some extent, yet the attacks generate new points out of manifold. The model now has to align these two distributions and it is forced to smooth the two energies to keep classifying both correctly. Once $\Ex$ smoothness does not hold, the model is incapable of performing the alignment. $\Ex$ smoothness is also a desirable property of EBMs. Over the past few years, various strategies have emerged to enhance robustness, such as techniques that weight the training samples like MART~\cite{wang2019improving}, GAIR-RST~\cite{zhang2020geometry} or approaches focused on smoothing the weight loss landscape, AWP~\cite{wu2020adversarial}. Furthermore, recent SOTA~\cite{cui2023decoupled,wang2023better} leverage synthesized data to increase robustness even further. Upon analyzing the distributions of $\Delta\Ex$ and $\Delta\Exy$ for all test samples, we observed that as the model's robustness increases, the energy distribution tends to approach zero, as depicted in~\cref{fig:energy_smoothness}. From the figure is also clear that the smoothing effect of TRADES compared to SAT is also visible in~\cref{fig:overfit}.

\subsection{AT with Delta Energy Regularization (DER)}
Our analysis of energy dynamics during AT reveals a consistent pattern: the norm  $\norm{[\Delta \Ex, \Delta \Exy]}_2$ increases significantly for certain samples in the presence of overfitting. This spike is particularly evident for AAEs during CO, see \cref{fig:abnormal_quiver} (RS-FGSM) in single-step AT, and for low-energy samples during RO under multi-step AT, as seen in \cref{fig:fig_quiver} (SAT). We therefore introduce Delta Energy Regularization (DER), penalizing samples for excessively large energy shifts:
\[
\mathcal{L}_{\Delta E_{\net}} = \max  \Big(\norm{  [\Delta \Ex, \Delta \Exy] }_2 - \gamma, 0 \Big).
\] 
\noindent The hyperparameter $\gamma$ allows flexibility by penalizing large deviations, while preventing penalties for smaller changes. In the single-step setting, the regularizer is selectively applied to AAEs. For multi-step AT, we observe that samples with lower energy tend to exhibit higher $\norm{(\Delta \Ex, \Delta \Exy)}_2$ values as evident from \cref{fig:binsDeltaEx}. Thus, our method implicitly penalizes such samples more heavily, without explicitly relying on an energy-based weighting scheme such as WEAT~\cite{mirza2024shedding}.

\begin{equation}
\begin{aligned}
\text{Single-step AT} &:\quad \mathcal{L}_{\text{total}} = \mathcal{L}_{\text{CE}}(\bxa) + \beta \cdot \mathds{1}_{\text{AAE}} \cdot \mathcal{L}_{\Delta E_{\net}} \\
\text{Multi-step AT} &:\quad \mathcal{L}_{\text{total}} = \mathcal{L}_{\text{CE}}(\bxa) + \beta \cdot \mathcal{L}_{\Delta E_{\net}}
\end{aligned}
\label{eq:DER} 
\end{equation}

\begin{figure*}[t]
    \centering
    \begin{minipage}[b]{0.5\textwidth}
        \centering
        \begin{overpic}[width=\linewidth, trim=2cm 0 2cm 0, clip]{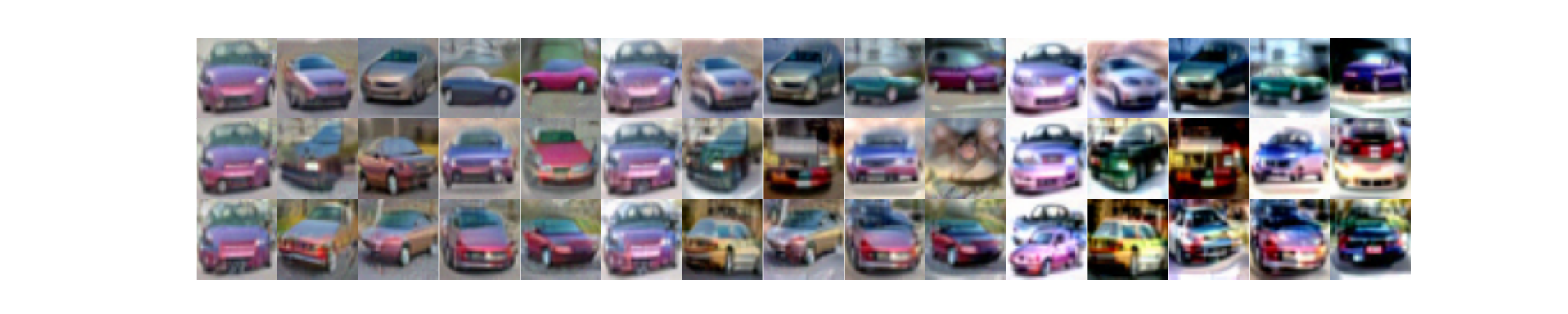}
            \put(36.8,2.5){\color{black}\linethickness{0.7pt}\line(0,1){17.1}}
            \put(65.9,2.5){\color{black}\linethickness{0.7pt}\line(0,1){17.1}}
            \put(0,19.8){\vector(0,-1){17}}
            \put(1.5,3.5){\rotatebox{90}{\scriptsize\shortstack{PCA retained \\ variance}}}
        \end{overpic}
        \vspace{0.3cm}
        \begin{overpic}[width=\linewidth, trim=2cm 0 2cm 0, clip]{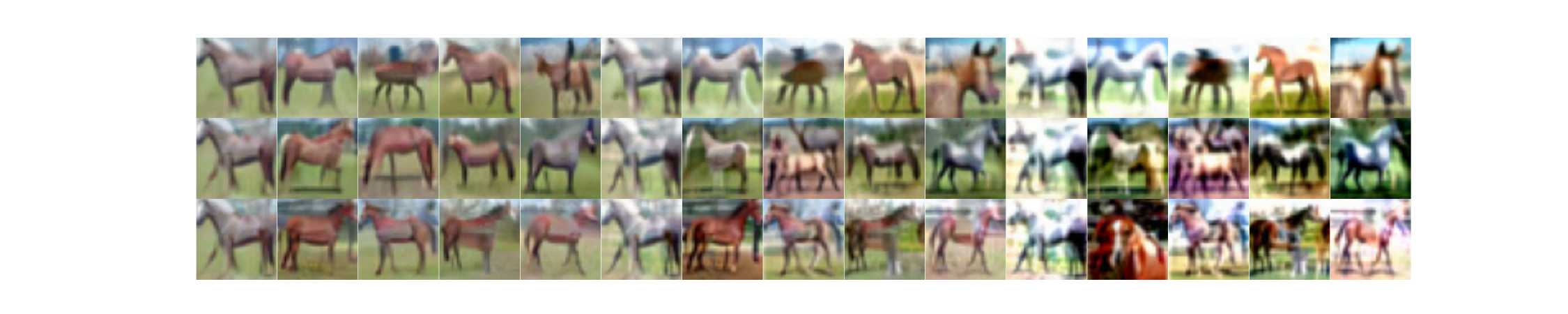}
            \put(36.8,2.5){\color{black}\linethickness{0.7pt}\line(0,1){17.1}}
            \put(65.9,2.5){\color{black}\linethickness{0.7pt}\line(0,1){17.1}}
            \put(0,19.8){\vector(0,-1){17}}
            \put(1.5,3.5){\rotatebox{90}{\scriptsize\shortstack{PCA retained \\ variance}}}
            \put(13.3,43.5){\tiny{\shortstack{ $\sigma_{PCA} = 0.005$ }}}
            \put(42.3,43.5){\tiny{\shortstack{ $\sigma_{PCA} = 0.01$ }}}
            \put(70.8,43.5){\tiny{\shortstack{ $\sigma_{PCA} = 0.02$ }}}
            \put(10,0){\vector(1,0){82}}
            \put(15,-3.5){{\shortstack{\small {Increasing $\sigma_{PCA}$ values for the generation}}}}
        \end{overpic}
        
    \end{minipage}
    \hfill
    \begin{minipage}[b]{0.49\textwidth}
           \centering
    \begin{subfigure}[b]{0.48\linewidth}
        \begin{overpic}[width=\linewidth]{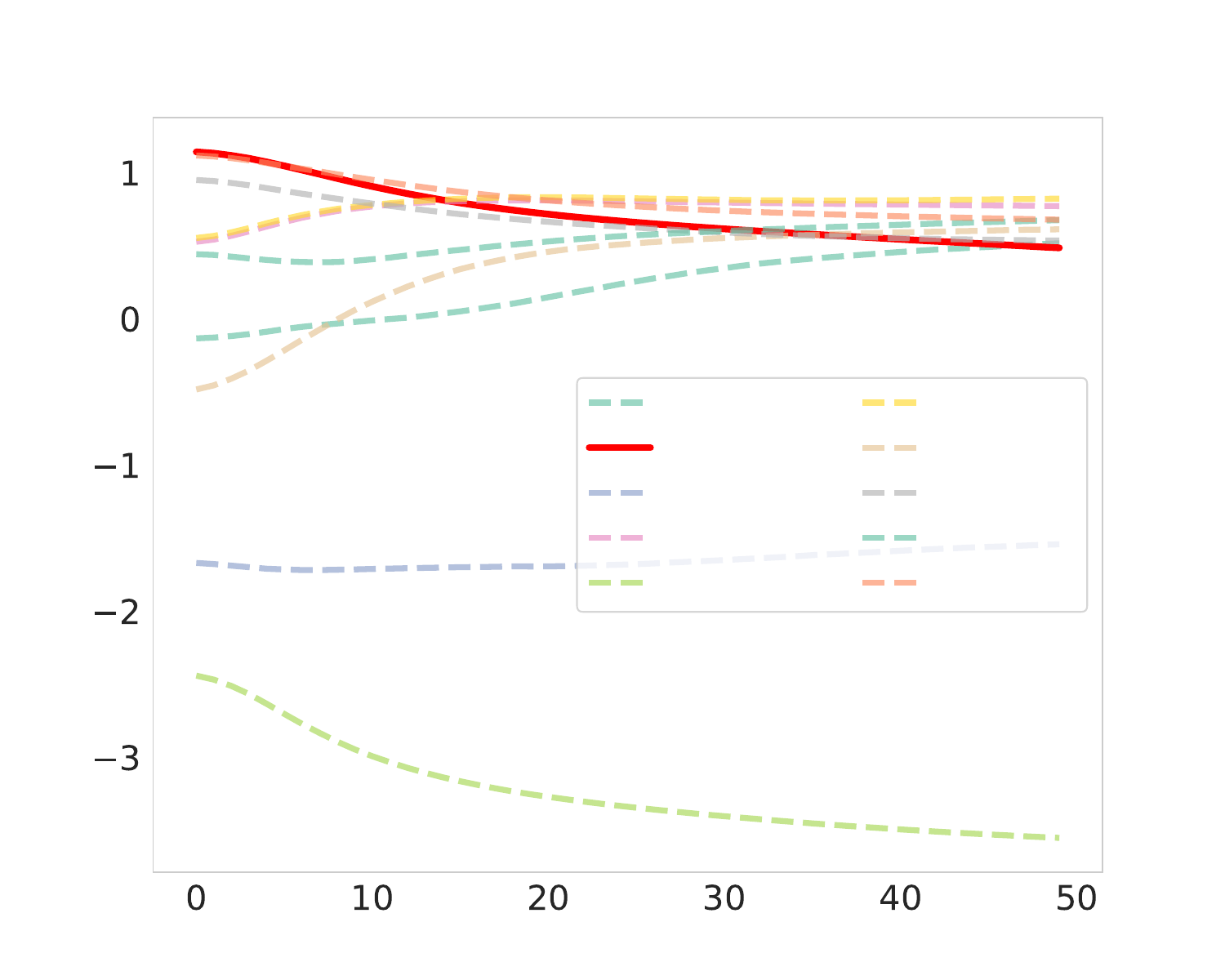}
            \put(35, -3){\small Iterations} 
            \put(-1, 25){\rotatebox{90}{\small $E(x,y)$}} 
            \put(54,46.5){\scalebox{.3}{$E(x,y_1)$}}
           \put(54,42.5){\scalebox{.3}{$E(x,y_2=\bar y)$}}
           \put(54,39){\scalebox{.3}{$E(x,y_1)$}}
           \put(54,35.5){\scalebox{.3}{$E(x,y_1)$}}
            \put(54,32){\scalebox{.3}{$E(x,y_1)$}}
            \put(76,46.5){\scalebox{.3}{$E(x,y_1)$}}
           \put(76,42.5){\scalebox{.3}{$E(x,y_1)$}}
           \put(76,39){\scalebox{.3}{$E(x,y_1)$}}
           \put(76,35.5){\scalebox{.3}{$E(x,y_1)$}}
            \put(76,32){\scalebox{.3}{$E(x,y_1)$}}           
        \end{overpic}
        \caption{}
        \label{subfig:wrong-EP}
    \end{subfigure}
    \hfill
    \begin{subfigure}[b]{0.48\linewidth}
        \begin{overpic}[width=\linewidth]{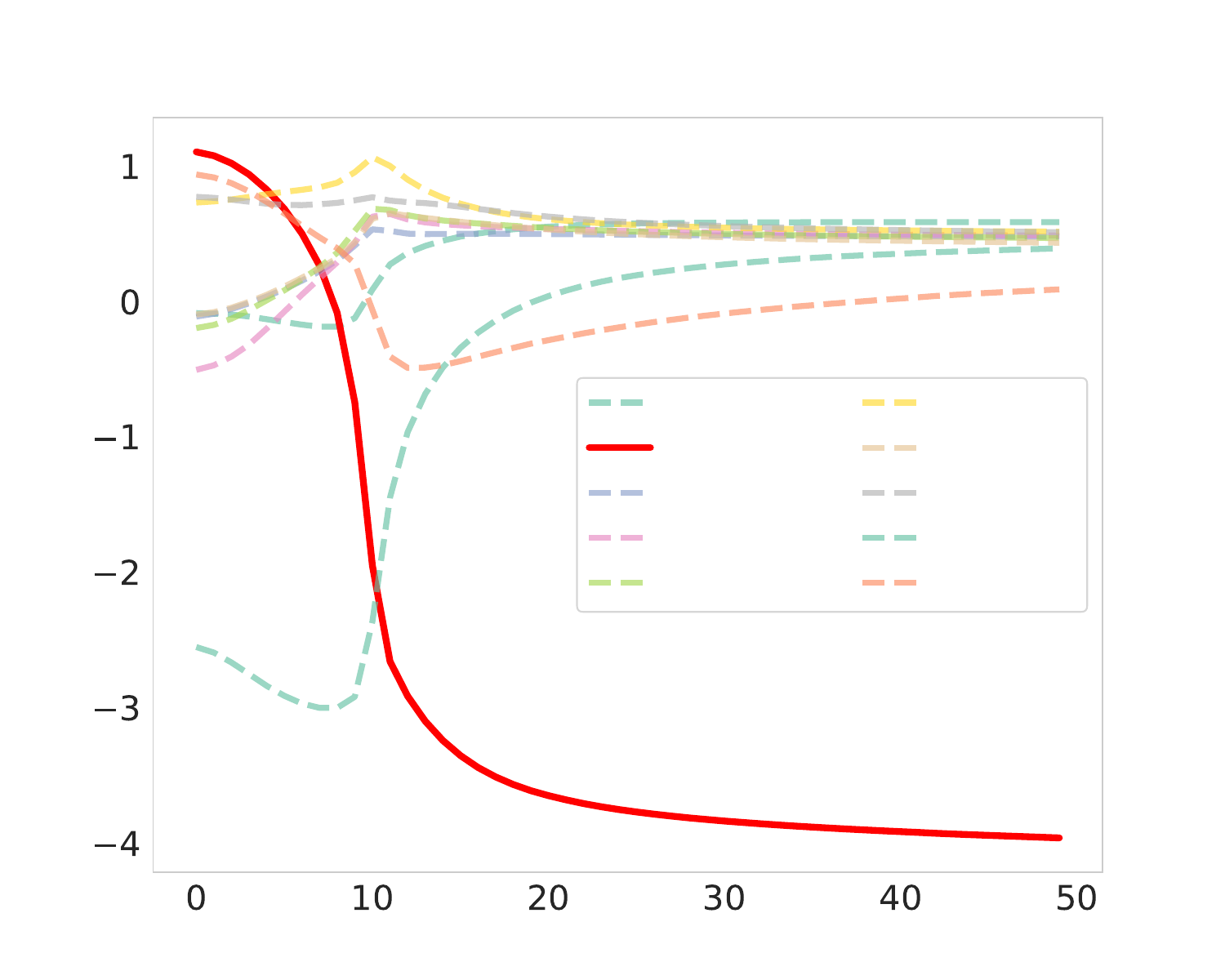}
            \put(35, -3){\small Iterations} 
            \put(-1, 25){\rotatebox{90}{\small $E(x,y)$}} 
            \put(54,46.5){\scalebox{.3}{$E(x,y_1)$}}
           \put(54,42.5){\scalebox{.3}{$E(x,y_2=\bar y)$}}
           \put(54,39){\scalebox{.3}{$E(x,y_1)$}}
           \put(54,35.5){\scalebox{.3}{$E(x,y_1)$}}
            \put(54,32){\scalebox{.3}{$E(x,y_1)$}}
            \put(76,46.5){\scalebox{.3}{$E(x,y_1)$}}
           \put(76,42.5){\scalebox{.3}{$E(x,y_1)$}}
           \put(76,39){\scalebox{.3}{$E(x,y_1)$}}
           \put(76,35.5){\scalebox{.3}{$E(x,y_1)$}}
            \put(76,32){\scalebox{.3}{$E(x,y_1)$}}     
        \end{overpic}
        \caption{}
        \label{subfig:EP}
    \end{subfigure}
    
\end{minipage}

    \caption{
        Left: Comparison of different PCA retained variances and $\sigma_{\text{PCA}}$ values. Rows correspond to $90\%$, $95\%$, and $99\%$ retained variance; columns show $\sigma_{\text{PCA}}=0.005$, $0.01$, and $0.02$. Right: Joint energy trends over 50-step generation when the sample is being generated for class $\bar y = y_2$ and samples in the starting  $K_\text{NN}$  are \textbf{(a)}  unfiltered by class or \textbf{(b)} filtered by $\bar{y}$.
    }
    \label{fig:combined-layout}
\end{figure*}
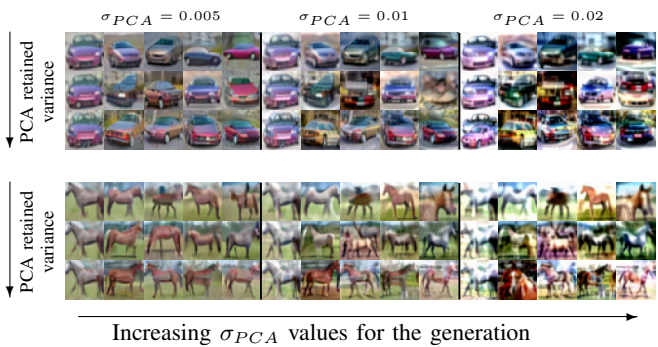

\section{Energy Impact on the Generation} 
Previous works~\cite{grathwohl2019your,zhu2021towards,wang2022aunified,mirza2024shedding} have extensively explored the generative capabilities of robust classifiers, aiming to develop frameworks for generating natural-looking images.
In particular, \cite{mirza2024shedding} shows the connection between classifier energy during the generation process and their training dynamics, highlighting how different training objectives can lead to significantly different generative behaviors.
It also introduces a PCA-based class-wise initialization followed by SGLD updates for image synthesis.
We found this approach limited in diversity due to the class-wise PCA initialization.

\begin{figure*}[ht]
    \centering
    \begin{subfigure}[b]{0.47\textwidth}
        \includegraphics[width=\textwidth]{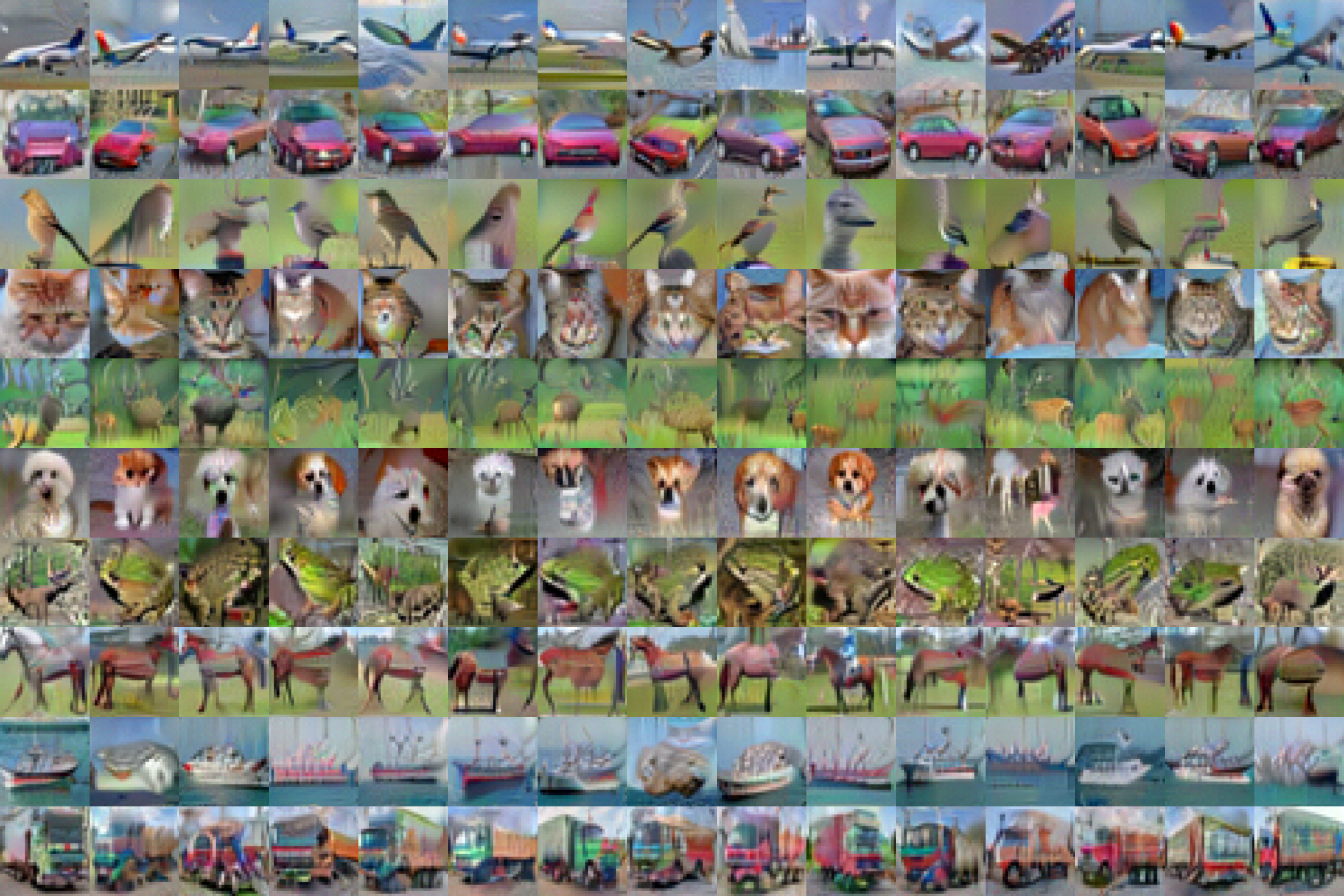}
        \caption{Per-class samples generated with \cite{mirza2024shedding}}
    \end{subfigure}
    \hspace{0.03\textwidth}
    \begin{subfigure}[b]{0.47\textwidth}
        \includegraphics[width=\textwidth]{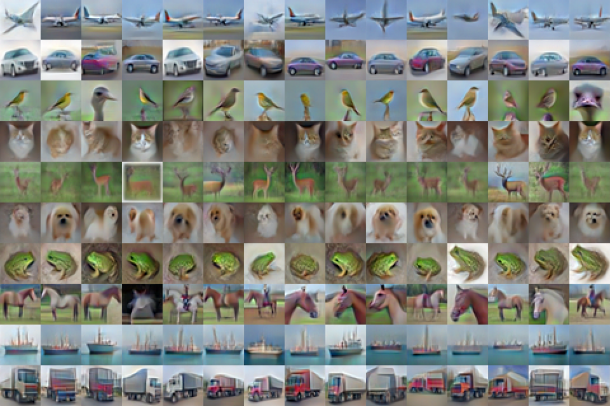}
        \caption{Per-class samples generated with our method}
    \end{subfigure}
    \caption{The figures display generations from the reference method \cite{mirza2024shedding} (left) and our proposed approach (right), both using the baseline method \cite{wang2023better} as baseline. In the left figure, the intra-class variability is notably limited, with images of the same class exhibiting highly similar attributes (e.g., predominantly red cars, identical airplane and bird backgrounds). In contrast, our method (right) demonstrates significantly greater intra-class variability, enhancing the diversity of generated samples. This improvement in variability comes at the cost of a slight reduction in image quality, losing fine details in some samples.}
    \label{fig:qualitative-comparison}
\end{figure*}
\subsection{Trade-Off between Quality and Diversity}
Upon examining the images shown in~\cite{mirza2024shedding}, it is evident that there is a bias in generations. For example, in the case of the \emph{car} class, a significant portion of the samples produced share common qualitative attributes (e.g., red color).
This bias is likely a consequence of the initial random sampling along the principal components: the attribute ``a red car'' is probably one of the strongest variations in the data, and the sampling method reflects this, heavily biasing the network features and consequently the generation. 
This hypothesis is then consistent with what was observed when the generation varies the retained variance of PCA and the parameters $\sigma_{\text{PCA}}$, the noise applied during the initialization sampling. 
In \cref{fig:combined-layout} (\textit{left}), there is a representation of the impact of parameters in the generation.
The main observation is the existence of a \textit{weak trade-off between the variability of the data and their quality}.
First, decreasing the explained variance results in images with less intricate details, owing to the reduced representation of informative features from the original image, but smoother, natural-looking images, keen on having common features among them, in particular regarding most discriminant class features (e.g., shape, pose, color, etc.). The same effect can be achieved by increasing the variance of the sampling noise.   
Manipulation of $\sigma_{\text{PCA}}$ introduces additional noise during initialization, leading to more variable generations, yet with diminished image quality, which will tend to preserve noise.

\subsection{Local Subspace Estimation allows Diversity in Synthesis}
In this work, we adopt the initialization method proposed in~\cite{mirza2024shedding}, exploiting its initialization that poses starting samples significantly closer to the real data manifold.  
Being those derived from a class-wise PCA algorithm, they inherently encode valuable information to be exploited during generation, leading to improved qualitative generations.
We propose an improved initialization technique based on a local class-wise PCA.
This algorithm allows for estimating the data manifold locally, fitting a local subspace through a sample neighborhood, with the aim at producing natural samples and meanwhile improving data variability.
At first, the initial $\mathbf{x}^s_0$ is a training data point belonging to the desired class $\bar y$.
Based on this, a set of $K$ training samples is selected as the top k most similar images with the highest SSIM \cite{Wang2004SSIM} w.r.t. $\mathbf{x}^s_0$.
Once the $K_\text{NN}$ set is defined, PCA is performed on it. 
This initialization method offers several advantages:
(i) Selecting the top $k$ most similar images ensures the consistency of the feature and appearance, resulting in a less noisy starting image; 
(ii) Since the starting point changes for each sample, the overall variability in the generated outputs will be improved, as different starting $K_\text{NN}$ sets lead to diverse initializations; 
PCA in this case is to be computed for each starting image, building $\textbf{x}_0$ from the $K_\text{NN}$ set.
Once defined $\textbf{x}_0$, SGLD algorithm is applied as in~\cite{mirza2024shedding} varying the objective function:
\begin{equation}
\begin{cases}
\mathcal{L}_\text{inv}(x)  = E(x,\bar y )- \phi\cdot E(x, \hat{y})~ \text{;}~ \hat{y} = \underset{y\neq \bar y }{\text{argmin}}~E(x, y)\\
    \bsf{\nu}_{n+1}=\zeta\bsf{\nu}_{n}-\frac{1}{2}\eta\nabla_{\bx} \mathcal{L}_\text{inv}(x) ~\text{;}~ \bsf{\nu}_{0}=\mbf{0}\\
    \bx_{n+1} = \bx_n + \bsf{\nu}_{n+1}  + \bsf{\varepsilon} ~\text{;}~ \bx_0 = \bsf{\mu}_{K_\text{NN}} + \sum_i \lambda_i\bsf{\alpha}_i \mbf{U}_{i}^{K_\text{NN}}
    \end{cases} 
\label{eq:sgld}
\end{equation}
where the stochasticity comes from the term $\bsf{\alpha}\sim \mathcal{N}(0;\sigma_\text{PCA})$. Here, $\bsf{\mu}_{K_\text{NN}}$ and $\mbf{U}^{K_\text{NN}}$ are the mean and the principal components of the $K_\text{NN}$ initialization set, $\phi$ is the weighting coefficient and $\lambda_i$ is the singular value associated to each component.
Moreover, SGLD is supposed to have random noise $\bsf{\varepsilon}\sim \mathcal{N}(0,\gamma\mbf{I})$,  then $\eta$ is the step size and $\zeta$ the friction coefficient. 
We use the same loss as in~\cite{zhu2021towards}, which is class dependent and allows us to sample from $p(\bx|y)$. By applying this loss in SGLD steps, we aim to decrease $E(x,\bar y )$ and increase joint energies w.r.t. to the other classes.

\subsection{The role of Energy during the generation}
By designing inference as described above, we observed that the energy plays a key role in guiding generation. Firstly, energy gives us clear hints on how to properly compose the $K_\text{NN}$ cluster. Suppose that we want to generate samples from class $\bar{y}$, a possible choice could be to select all elements from the entire data set to populate the starting cluster, without filtering them by class. In this case, observing the trend of the joint energies while generating in \cref{subfig:EP,subfig:wrong-EP}, we can see that the expected behavior of $E(\textbf{x},\bar{y})$ is pushed down until its average value per class does not happen. Energies wrt. other classes also decrease, even more than $E(\textbf{x},\bar{y})$, causing undesired features to appear in the generated samples.
To overcome this problem, we chose the $K_\text{NN}$ elements among the data of the desired class $\bar{y}$. In this case, the expected behavior always appears to be consistent with what was expected, properly lowering $E(\textbf{x},\bar{y})$ and consistently pushing up all the other joint energies. In \cref{fig:combined-layout} (\textit{right}), the red line is $E(\textbf{x}, \bar{y})$, which is being minimized in the generation procedure. In \cref{subfig:wrong-EP} we can see that $E(\textbf{x}, \bar{y})$ is not correctly lowered, whereas energies wrt. other classes are. Instead, in \cref{subfig:EP} we can observe that $E(\textbf{x}, \bar{y})$ is correctly pushed down, meanwhile joint energies wrt. other classes are pushed up. Beyond guiding the selection of an appropriate initialization cluster, energy also plays a key role in the generation dynamics. Unlike~\cite{mirza2024shedding}, the generation of $n$ images requires $n$ distinct $\bx_0$, calculated from the $K_\text{NN}$ sets. Each point has a different initial energy $E(\textbf{x}_0, y)$ wrt. each dataset class. This variability must be taken into account during inference, since samples' generation trajectories can differ significantly, making it challenging to define unique hyperparameters. To address this, energy is used as a guide for image generation. As noted in~\cite{mirza2024shedding},  samples from class $\bar{y}$ can be considered fully generated once their energies converge to the mean of that class, calculated employing the identical classifier utilized in the generation process. This insight allows for avoiding setting a fixed number of iterations, introducing instead a stopping criterion that sample-wise varies the number of iterations, but still ensuring the generation of natural-looking images across different initialization clusters.
\begin{table*}[ht]
    \centering
    \begin{minipage}{0.85\linewidth}
        \resizebox{\linewidth}{!}{
        \setlength{\tabcolsep}{3pt} 
    \begin{tabular}{c@{\hskip 5pt}ccc@{\hskip 20pt}ccc@{\hskip 20pt}ccc@{\hskip 20pt}ccc}
    \toprule
    \multirow{2}{*}{\renewcommand{\arraystretch}{0.9}%
    \begin{tabular}{@{}c@{}}Defence\\method\end{tabular}} 
    & \multicolumn{3}{c@{\hskip 20pt}}{CIFAR-10} 
    & \multicolumn{3}{c@{\hskip 20pt}}{CIFAR-100} 
    & \multicolumn{3}{c@{\hskip 20pt}}{SVHN} 
    & \multicolumn{3}{c}{Tiny-ImageNet} \\
    \cmidrule(lr){2-4} \cmidrule(lr){5-7} \cmidrule(lr){8-10} \cmidrule(lr){11-13} 
    & Clean & PGD & AA & Clean & PGD & AA & Clean & PGD & AA & Clean & PGD & AA \\
    \midrule
    {\textsc{rs-fgsm}~\cite{wong2020f}} & 89.48 & 00.00 & 00.00 & 46.90 & 0.00 & 0.00 & 93.9 & 0.00 & 0.00 & 48.80 & 0.00 & 0.00 \\
    {\textsc{n-fgsm}~\cite{de2022make}} & 82.24 & 48.12 & 44.57 & 57.24 & 26.01 & 22.64 & \textbf{91.57} & 43.25 & 36.15 & 49.98 & 20.41 & \textbf{17.94} \\
    {\textsc{dom$_{\text{re}}$}*\cite{lin2024on}} & 71.66 & 47.09 & 17.10 & 26.39 & 12.68 & 7.65 & 89.96 & \textbf{47.92} & 32.22 & --- & --- & --- \\
    {\textsc{dom$_{\text{da}}$}*\cite{lin2024on}} & 84.52 & \textbf{50.10} & 42.53 & 55.09  & \textbf{27.44} & 21.38 & 85.88 & 51.57 & 36.69 & --- & --- & --- \\
    \textsc{rs-aaer~\cite{lin2024eliminating}} & \textbf{85.60} & 45.88 & 42.54 & \textbf{59.03} & 24.62 & 20.52 & 0.06 & 0.06 & 0.00 & \textbf{52.02} & 19.24 & 16.35 \\
    \textsc{n-aaer~\cite{lin2024eliminating}} & 82.50 & 48.11 & 44.36 & 57.15 & 25.72 & 22.33 & 90.81 & 46.37 & \textbf{40.50} & 49.61 & 20.57 & 17.23 \\
    \textsc{rs-der} (Ours) & 85.45 & 45.45 & 42.88 & 55.61 & 24.77 &  21.29 & 87.53 & 38.67 & 32.51 & 44.59 & 18.23 & 14.28 \\
    \textsc{n-der} (Ours)& 82.45 & 48.30 & \textbf{44.60} & 58.89 & 26.26 &  \textbf{23.07} & 90.34 & 45.90 & 39.48 & 48.31 & \textbf{21.29} & {17.51} \\
    \bottomrule
\end{tabular}
        }
        \caption{Single-step AT results using PreActResNet-18 (\(\epsilon = 8/255\)), evaluated with PGD-20 and AutoAttack (AA).{\footnotesize \textit{ Results marked with an asterisk (*) are taken from their original papers.}}} 
        \label{tab:single_step_eps_8}
    \end{minipage}%
\end{table*}

\begin{table}[t]
\centering
\setlength{\tabcolsep}{2pt}
\begin{tabular}{lcc@{\hskip 10pt}cc@{\hskip 10pt}cc@{\hskip 10pt}cc}
\toprule
\multirow{2}{*}{Method} 
& \multicolumn{4}{c@{\hskip 10pt}}{CIFAR-10} 
& \multicolumn{4}{c}{CIFAR-100} \\
\cmidrule(lr){2-5} \cmidrule(lr){6-9}
& \multicolumn{2}{c}{16/255} & \multicolumn{2}{c@{\hskip 10pt}}{32/255}
& \multicolumn{2}{c}{16/255} & \multicolumn{2}{c}{32/255} \\
\cmidrule(lr){2-3} \cmidrule(lr){4-5}
\cmidrule(lr){6-7} \cmidrule(lr){8-9}
& Clean & PGD & Clean & PGD 
& Clean & PGD & Clean & PGD \\
\midrule
\textsc{rs-trades}*~\cite{zhang2019theoretically} & 91.50 & 0.01 & 88.69 & 0.00 & --- & --- & --- & --- \\
\textsc{n-trades}*~\cite{zhang2019theoretically} & 68.16 & 26.13 & 85.97 & 0.24 & --- & --- & --- & --- \\
\textsc{rs-alp}*~\cite{kannan2018adversarial} & 90.48 & 0.00 & 81.03 & 0.00 & --- & --- & --- & --- \\
\textsc{n-alp}*~\cite{kannan2018adversarial} & \textbf{69.32} & 23.46 & 82.98 & 0.00 & --- & --- & --- & --- \\
\textsc{nuat}*~\cite{sriramanan2021towards} & 80.10 & 3.29 & --- & --- & --- & --- & --- & --- \\
\textsc{rs-fgsm}*~\cite{wong2020f} & 66.54 & 0.00 & 36.43 & 0.00 & 11.03 & 0.00 & 11.40 & 0.00 \\
\textsc{n-fgsm}*~\cite{de2022make} & 62.96 & 27.14 & 29.79 & 8.30 & \textbf{37.93} & 14.05 & 18.18 & 0.00 \\
\textsc{rs-aaer*~\cite{lin2024eliminating}} & 64.56 & 23.87 & 31.58 & 10.62 & 33.10 & 11.80  & \textbf{18.50} & 4.90 \\
\textsc{n-aaer*~\cite{lin2024eliminating}} & 61.84 & 28.20 & 27.08 & 12.97 & 36.80 & \textbf{14.31}  & 16.95 & \textbf{5.45} \\
\textsc{rs-der} (Ours) & 49.79 & 22.32 & 28.60 & \textbf{13.88} & 28.82 & 10.61 & 15.05  & 4.79 \\
\textsc{n-der} (Ours)& 62.21 & \textbf{28.36} & \textbf{33.23} & 10.19 & 36.32 & 14.01 & 17.93  & 4.63 \\
\midrule
\textsc{SAT}*~\cite{madry2017towards}$\mathrlap{\text{\tiny~(multi-step)}}$& 67.20 & {29.34} & {34.70} & 16.10 & 42.39 & 15.01 & 21.68  & 7.39 \\

\bottomrule
\end{tabular}
\caption{Single-step AT results with PreActResNet-18 under \(\epsilon = 16/255\) and \(32/255\), evaluated with PGD-50-10.\\
{\footnotesize \textit{ Results marked with an asterisk (*) are taken from~\cite{lin2024eliminating}.}}}
\label{tab:single_step_eps_16_32}
\end{table}

\section{Experimental Evaluation}\label{sec:expts}
Our experiments are structured to offer deeper insights into overfitting in adversarial training (AT) by leveraging an energy-based framework, rather than focusing solely on surpassing existing SOTA methods. We present comprehensive evaluations that not only demonstrate the robustness of our proposed method but also reinterpret existing SOTA approaches in terms of energy dynamics.
To provide a thorough understanding, we assess the robustness of our proposed method, validate design choices and finally evaluate the generative quality of images produced by different SOTA robust classifiers. We quantitatively assess image quality and diversity using established metrics like IS~\cite{salimans2016improved} and  FID~\cite{heusel2017gans}, evaluated on $50,000$ images. We aim to illustrate that the new initialization method enables diversity in the generation while still getting comparable results in the evaluation metrics w.r.t. to~\cite{mirza2024shedding}, while improving sample variability.

\subsection{Implementation Details}
We perform AT under the \(L_{\infty}\) threat model on four standard benchmark datasets: CIFAR-10~\cite{krizhevsky2009learning}, CIFAR-100~\cite{krizhevsky2009learning}, SVHN~\cite{yuval2011reading} and Tiny-ImageNet~\cite{deng2009imagenet}.
In single-step AT setting, the methods RS-FGSM~\cite{wong2020f}, N-FGSM~\cite{de2022make} and AAER~\cite{lin2024eliminating}, were retrained from scratch using their publicly available code. To ensure a fair comparison, all methods were trained using the same architecture, PreActResNet-18 and  evaluation is done on the final checkpoint. The models are trained on CIFAR-10/100 and Tiny-ImageNet for 100 epochs in the multi-step setting, and for 50 epochs in the single-step setting. For SVHN, all models are trained for 30 epochs in both settings. While single-step training is typically conducted with fewer epochs, we extend it to 50 epochs to examine model behavior under longer training regimes. For single-step training, we follow the settings used in~\cite{lin2024eliminating}, while for multi-step training, we adopt the configuration from~\cite{mirza2024shedding}. Moreover, in multi-step AT, we initially train the model using SAT~\cite{madry2017towards} and apply the regularizer during the final epochs, preferably starting at the onset of RO. For RS-DER, $\beta{=}0.5$ ($\epsilon{=}8/255$), and $4$ (CIFAR-10) / $1.5$ (CIFAR-100) at higher $\epsilon$. N-DER uses $\beta{=}0.1$ across all settings. In multi-step, DER has $\beta{=}$6 (CIFAR-10), 3 (CIFAR-100, TinyImageNet), 1 (SVHN); $\gamma{=}0.2$, except for SVHN in single-step setting where $\gamma{=}0$. For robustness, we evaluate with PGD-20 (20 steps) or PGD-50-10 (50 steps with 10 random restarts), both with step size \(\alpha = \epsilon/4\), and AA.\\
We use CIFAR-10 to assess the generative capabilities of various SOTA robust classifiers from RobustBench. While generation, we preserve $99\%$ of data variance, effectively guaranteeing a certain amount of starting information while minimizing high-frequency noise. Parameters such as friction $\zeta$, noise variance $\gamma$, and step size $\eta$ are set to $0.8$, $0.001$, and $0.05$ respectively. The number of SGLD steps ($N$) for each sample is dynamically set. Given $\bar X = \{x\} ~\forall (x,y) \in \mathcal D~  | ~ y = \bar y$, the set of samples belonging to the class $\bar y$, we define $\mu_{E(x,\bar y)} = \frac{1}{|\bar X|} \underset{\bar X}{\sum }E(x, \bar y)$ and $\sigma_{E(x, \bar y)} = \sqrt{ \frac{1}{|\bar X|} \sum_{x \in \bar X} \left(E(x, \bar y) - \mu_{E(x, \bar y)}\right)^2 }$, the generation loops end as soon as $E(x_i,\bar y) < (\mu _{E(x, \bar y)} - \sigma_{E(x, \bar y)})$.  With these choices, energy descent stays smooth over the generation steps, where images are projected to the range $[0, 1]$ at each iteration.
\color{black}

\begin{table}[ht]
\centering
\setlength{\tabcolsep}{3pt} 
\renewcommand{\arraystretch}{0.90} 
    \begin{tabular}{lcccc@{\hskip 10pt}ccc}
    \toprule
    \multirow{2}{*}{\centering Defense} & Dataset & \multicolumn{3}{c@{\hskip 10pt}}{Best} & \multicolumn{3}{c}{Last} \\
    \cmidrule(lr){3-5} \cmidrule(lr){6-8}
    & & Nat & PGD & AA & Nat & PGD & AA \\
    \midrule
    \multirow{4}{*}{\centering \shortstack{   \textsc{     trades} \\ \cite{zhang2019theoretically}}}  
    & CIFAR-10 & 82.91 & 52.65 & 49.46 & 83.01 & 52.39 & 49.19 \\  
    & CIFAR-100 & 56.31 & 28.53 & 24.29 & 56.37 & 28.23 & 24.09 \\  
    & SVHN & 89.09 & 55.52 & 48.13 & 89.09 & 55.52 & 48.13 \\  
    & Tiny-ImageNet & 48.34 & 21.73 & 16.90 & 48.20 & 21.51 & 16.95 \\   
    \midrule
    \multirow{4}{*}{\centering \shortstack{\textsc{mail-tr} \\ \cite{liu2021probabilistic}}}  
    & CIFAR-10 & 81.63 & 53.09 & 49.42 & 82.14 & 52.56 & 49.11 \\  
    & CIFAR-100 & 56.30 & 28.79 & 24.24 & 56.45 & 28.52 & 24.16 \\  
    & SVHN & 89.65 & 54.94 & 47.48 & 
    89.65 & 54.94 & 47.48 \\  
    & Tiny-ImageNet & 48.88 & 21.99 & 17.23 & 48.67 & 21.96 & 16.90 \\ 
    \midrule
    \multirow{4}{*}{\centering \shortstack{\textsc{weat$_{\text{adv}}$} \\ \cite{mirza2024shedding}}}  
    & CIFAR-10 & 80.89 & 53.11 & \textbf{49.85} & 80.89 & 53.11 & \textbf{49.85} \\  
    & CIFAR-100 & 54.49 & 29.05 & 24.26 & 54.49 & 29.05 & 24.26 \\  
    & SVHN & 86.90  & \textbf{56.68} & \textbf{49.93} & 86.90  & \textbf{56.68}  & \textbf{49.93} \\  
    & Tiny-ImageNet & 46.57 & 22.06 & 17.03 & 45.89 & 21.90 & 16.99 \\ 
    \midrule
    \multirow{4}{*}{\centering \shortstack{\textsc{~~~sat} \\~~~ \cite{madry2017towards}}}  
    & CIFAR-10 & 82.43 & 49.03 & 45.37 & \textbf{85.17} & 47.07 & 44.47 \\ 
    & CIFAR-100 & 54.78 & 23.89 & 20.99 & \textbf{58.49} & 22.63 & 20.81 \\ 
    & SVHN & \textbf{93.22} & 50.54 & 44.87 &  \textbf{93.22} & 50.54 & 44.87 \\  
    & Tiny-ImageNet & 48.91 & 20.54 & 16.87 & \textbf{51.38} & 19.01 & 16.23 \\  
    \midrule
     \multirow{4}{*}{\centering \shortstack{\textsc{mail-at} \\ \cite{liu2021probabilistic}}} 
    & CIFAR-10 & \textbf{83.77} & 53.36 & 42.60 & 84.42 & 52.97 & 42.89 \\  
    & CIFAR-100 & \textbf{56.90} & 25.39 & 19.6 & 57.22 & 25.28 & 19.59 \\  
    & SVHN & 92.97 & 52.63 & 41.7 & 
     92.97 & 52.63 & 41.70\\  
    & Tiny-ImageNet & \textbf{50.35} & 21.69 & 15.72 & 50.36 & 21.60 & 15.89 \\ 
    \midrule
    \multirow{4}{*}{ \centering \shortstack{\textsc{der-at} \\ {(Ours)}}}
    & CIFAR-10 & 82.01 & \textbf{55.02} & 47.67 & 82.57 & \textbf{54.95} & 47.91 \\  
    & CIFAR-100 & 53.93  & \textbf{30.47} & \textbf{24.50}  & 54.47   & \textbf{30.15} & \textbf{24.38} \\  
    & SVHN & 90.68 & 54.67 & 47.31 & 90.68 & 54.67 & 47.31 \\  
    & Tiny-ImageNet & 46.41 & \textbf{25.24} & \textbf{18.01} & 46.62 & \textbf{24.80} & \textbf{17.76} \\  
    \bottomrule
\end{tabular}
\begin{tikzpicture}[remember picture, overlay]

\draw[decorate,decoration={brace,amplitude=4pt,mirror},xshift=-8.7cm,yshift=3.5cm]
  (0,0) -- (0,-3.8) node[midway,xshift=-0.3cm,rotate=90]{\small Inner max. $\mathcal{L}_\text{~KL-Divergence}$};

\draw[decorate,decoration={brace,amplitude=4pt,mirror},xshift=-8.7cm,yshift= -0.5cm]
  (0,0) -- (0,-3.8) node[midway,xshift=-0.3cm,rotate=90]{\small Inner max. $\mathcal{L}_\text{~Cross-Entropy}$};
\end{tikzpicture}
\caption{Multi-step AT with ResNet-18: PGD-20 and AutoAttack (AA) results for best and last checkpoint (\(\epsilon=8/255\)).}
\label{tab:multi_step_eps_8}
\end{table}

\subsection{Results on Adversarial Training}
We report the mean of the results from three models trained with different random seeds; standard deviations are less than 0.4 and omitted for the sake of clarity. The single-step results under the $L_\infty$ threat model are shown in \cref{tab:single_step_eps_8} ($\epsilon = 8/255$) and \cref{tab:single_step_eps_16_32} ($\epsilon = 16/255, 32/255$). At $\epsilon = 8/255$ with longer training, RS-FGSM suffers from CO, whereas applying AAER or DER to RS-FGSM entirely prevents CO and maintains both clean and adversarial accuracy. However, on the SVHN dataset, RS-AAER suffers from CO or unstable training. At larger perturbations, with the exception of AAER and our proposed DER, all baseline methods—RS-FGSM, N-FGSM, and other SOTA defenses such as NuAT~\cite{sriramanan2021towards}, ALP~\cite{kannan2018adversarial} and TRADES~\cite{zhang2019theoretically} exhibit degraded performance or succumb to CO. While AAER relies on complex regularization terms and multiple hyperparameters—sometimes leading to instability during training—our regularizer adopts a much simpler design with fewer hyperparameters, primarily relying on $\beta$. In the multi-step setting, we aim to demonstrate that RO, which is particularly prominent when training with SAT~\cite{madry2017towards}, can be effectively mitigated with minimal intervention. As evidenced in \cref{tab:multi_step_eps_8}, applying our regularizer, DER, only during the final phases of training is sufficient to prevent RO and achieve high robust accuracy. While related methods such as MLCAT with logit scaling~\cite{yu2022understanding} and MAIL-AT~\cite{liu2021probabilistic} also improve robustness against untargeted-PGD attack, they often do so at the cost of reduced robustness to AA. In contrast, our approach not only avoids such degradation but also improves robustness against AA. We also conduct a broader comparison among defenses that use either KL divergence or CE loss as the inner maximization objective in \cref{eq:adversarial_training}, and observe that KL-based approaches typically demonstrate stronger robustness under AA. The key distinction lies in how these objectives guide adversarial example generation. CE focuses solely on increasing the loss for the ground-truth label, effectively reducing the model’s confidence in the correct class. In contrast, KL divergence leads the adversary to perturb not just the ground-truth prediction but also the relative confidence among all classes, weighted by the classifier’s predicted probabilities, as shown in \cref{prop:1}. Thus, approaches like TRADES~\cite{zhang2019theoretically}, which use KL divergence both for data perturbation and as a training regularizer, tend to perform slightly better than DER on targeted attacks such as APGD-T (part of AA), but generally show slightly lower robustness on untargeted PGD attacks. These benefits are more pronounced on well-modeled datasets like CIFAR-10 and SVHN, while on more challenging datasets such as CIFAR-100 and Tiny-ImageNet, the AA robustness of TRADES and DER becomes comparable. 

\minisection{Results under Extended Training for RO} To investigate the impact of prolonged training on RO, we extend the training schedule to 200 epochs, using an initial learning rate of 0.1. The learning rate decays by a factor of 10 at the 100th and 150th epochs, respectively. We show the results for DER with $\beta=8$ in \cref{tab:long_train}. We compare our approach with SOTA methods designed to reduce overfitting in adversarial training, including MLCAT~\cite{yu2022understanding}, KD~\cite{chen2020robust}, and DOM~\cite{lin2024on}. For a fair comparison, we evaluate against the loss-scaling variant of MLCAT and the smoothed-logits version of~\cite{chen2020robust}, excluding its extension with adversarial weight perturbation(AWP)~\cite{wu2020adversarial} and stochastic weight averaging~\cite{izmailov2018averaging}, which fall under parameter correction strategies. For \text{DOM$_{\text{DA}}$}, we consider the RandAugment variantas it also avoids CO in single-step training. While DOM, like our method, addresses both RO and CO, it does not assess CO under larger perturbation bounds in the single-step setting. Moreover, our approach also more effectively mitigates RO, particularly under the multi-step setting.

\begin{table}[t]
  \centering
  \scriptsize
  \renewcommand{\arraystretch}{1.1}  
  \setlength\tabcolsep{1.5pt}          
  \begin{tabularx}{\columnwidth}{
      @{} l
      *{5}{>{\centering\arraybackslash}X@{\hspace{1pt}}>{\centering\arraybackslash}X@{\hspace{6pt}}}
      @{}}
    \toprule
    & \multicolumn{2}{c}{\textsc{mlcat$_{\text{ls}}$}*~\cite{yu2022understanding}}
    & \multicolumn{2}{c}{\textsc{kd}*~\cite{chen2020robust}}
      & \multicolumn{2}{c}{\textsc{dom$_{\text{re}}$}*~\cite{lin2024on}}
      & \multicolumn{2}{c}{\textsc{dom$_{\text{da}}$}*~\cite{lin2024on}}
      & \multicolumn{2}{@{\hspace{-5pt}}c}{\textsc{der-at}$\mathrlap{\text{\tiny~(Ours)}}$} \\
    \cmidrule(lr){2-3}\cmidrule(lr){4-5}\cmidrule(lr){6-7}\cmidrule(lr){8-9}\cmidrule(lr){10-11}
    Metric & Best & Last & Best & Last & Best & Last & Best & Last & Best & Last \\
    \midrule
    Clean & ---     & ---   & \textbf{83.67} & \textbf{85.25}  & 80.23 & 80.66 & 83.49 & 83.74  & 82.07 & 83.29 \\
    PGD   & \textbf{56.90} & \textbf{56.87} & 50.89 & 48.26 & 55.48 & 52.52 & 52.83 & 50.39 &  55.17 & 54.17 \\
    AA    & 28.12 & 26.93 & --- & --- & 42.87 & 32.90 & \textbf{48.41} & 46.62 &  48.35 & \textbf{47.56} \\
    \midrule
    Clean & ---     & ---     & \textbf{57.42} & \textbf{60.34} &52.70 & 52.67 & 55.20 & 56.21 &  53.82 & 55.33 \\
    PGD   & 20.09 & 18.14 & 27.56 & 26.02 & 29.45 & 25.14 & 30.01 & 25.84  & \textbf{33.02} & \textbf{32.73} \\
    AA    & 13.41 & 11.35 & --- & --- & 20.41 & 17.59 & 24.10 & 20.79 &  \textbf{26.29} & \textbf{26.30} \\
    \bottomrule
  \end{tabularx}
  \begin{tikzpicture}[remember picture, overlay]
    \node[anchor=east, font=\tiny,rotate=270] at (4.40,0.1) {CIFAR-100};
    \node[anchor=east, font=\tiny,rotate=270] at (4.40,1.25) {CIFAR-10};
  \end{tikzpicture}
    \caption{Results on CIFAR-10 (top) and CIFAR-100 (bottom) using ResNet-18 under multi-step AT with \(\epsilon=8/255\), trained for \textbf{200 epochs}, evaluated using PGD-20 and AutoAttack (AA). {\footnotesize \textit{Results marked with an asterisk (*) are from original papers.}}}
  \label{tab:long_train}
\end{table}

\subsubsection{Ablation Study}\label{sec:ablation} We perform an ablation study to evaluate the influence of the hyperparameters in our DER regularizer~\cref{eq:DER}. As shown in \cref{tab:ablation_gamma}, setting $\gamma = 0.2$ leads to a modest improvement in clean accuracy without sacrificing robustness, compared to using $\gamma = 0$. Additionally, we observe that the choice of $\beta$ plays a key role in balancing clean and robust accuracy, highlighting its importance in controlling the regularization strength. Furthermore, we experimented with different regularizers as shown in~\cref{tab:ablation_multi_step}, such as KL divergence, or methods directly using logits of the model, such as ALP~\cite{kannan2018adversarial}. These methods are closely related to our approaches as they also influence the energy landscape. Specifically, KL divergence's effect on energies is demonstrated in \cref{prop:1}, whereas ALP can be viewed as minimizing the $\| \Exk - \Exkp \|_2^2$ for each class $k$, effectively aligning the joint energy distribution over all classes between clean and adversarial examples. Both the regularizers enhance the baseline SAT by improving both robustness against untargeted-PGD and AA.

\begin{table*}[ht]
\centering
\begin{subtable}[t]{0.78\linewidth}
\centering
\addtolength{\tabcolsep}{-0.52em}
\resizebox{0.99\linewidth}{!}{
\setlength{\tabcolsep}{2pt}
\begin{tabular}{l@{\hskip 5pt}ccc@{\hskip 10pt}ccc@{\hskip 10pt}ccc@{\hskip 10pt}ccc}
\toprule
\multirow{2}{*}{\textbf{Outer Loss Function}} 
& \multicolumn{6}{c@{\hskip 10pt}}{\textbf{CIFAR-10}} 
& \multicolumn{6}{c}{\textbf{CIFAR-100}} \\
\cmidrule(lr){2-7} \cmidrule(lr){8-13}
& \multicolumn{3}{c}{Best} & \multicolumn{3}{c@{\hskip 10pt}}{Last} 
& \multicolumn{3}{c}{Best} & \multicolumn{3}{c}{Last} \\
\cmidrule(lr){2-4} \cmidrule(lr){5-7} \cmidrule(lr){8-10} \cmidrule(lr){11-13}
& Clean & PGD & AA & Clean & PGD & AA 
& Clean & PGD & AA & Clean & PGD & AA \\
\cmidrule(lr){1-1}
$\mathcal{L}_{\text{CE}} + \lambda \| f(\bx) - f(\bxa) \|_2^2$
& 81.20 & 53.90 & 48.35 & 82.37 & 53.00 & 48.37 
& 55.56 & 28.48 & 24.81 & 55.82 & 28.21 & 24.52 \\
\cmidrule(lr){1-1}
$\mathcal{L}_{\text{CE}} + \lambda~ \mathcal{L}_{\text{KL}}$ 
& 79.72 & 53.25 & 47.72 & 82.22 & 52.22 & 47.30 
& 54.35 & 28.72 & 24.16 & 55.10 & 28.09 & 23.66 \\
\arrayrulecolor{gray}
\specialrule{0.025em}{0.4em}{0em}
\specialrule{0.025em}{0.05em}{0.4em}
\arrayrulecolor{black} 
$\frac{1}{n} \sum_i w_i \mathcal{L}_{\text{CE}}^{i}\mathrlap{\text{\tiny~(Not normalized)}}$
& 84.67 & 53.72 & 43.79 & 84.77 & 53.33 & 43.45 
& 57.76 & 31.91 & 17.00 & 57.64 & 31.74 & 16.95 \\
\cmidrule(lr){1-1}
$\frac{1}{\sum_i w_i} \sum_i w_i \mathcal{L}_{\text{CE}}^{i}\mathrlap{\text{\tiny~(Normalized)}}$
& 82.83 & 54.98 & 38.23 & 84.50 & 54.12 & 38.99
& 53.56 & 22.62 & 18.42 & 57.37 & 21.29 & 17.23 \\
\bottomrule
\end{tabular}
}
\caption{}
\label{tab:ablation_multi_step}
\end{subtable}
\hfill
\begin{subtable}[t]{0.21\linewidth}
\centering
\resizebox{0.99\linewidth}{!}{
\begin{tabular}{llcc} 
    \toprule
    $\boldsymbol{\gamma}$ & $\boldsymbol{\beta}$ & \begin{tabular}{@{}l@{}}Clean\\Acc.\end{tabular} & \begin{tabular}{@{}l@{}}APGD-\\~~CE\end{tabular} \\
    \midrule
    \multirow{4}{*}{0} & 
    2.5 & 82.99 & 52.33 \\ 
    & 3 & 82.47 & 52.78 \\
    & 3.5 & 82.25 & 53.02 \\
    & 4 & 81.31 & 53.14 \\
    \midrule
    \multirow{5}{*}{2} & 4 & {83.58} & 52.86 \\
    & 5 & 83.24 & 53.57 \\
    & 6 & 82.7 & 54.08 \\
    & 7 & 82.43 & 53.89 \\
    & 8 & 82.11 & {54.24} \\
    \bottomrule
\end{tabular}
}
\caption{}
\label{tab:ablation_gamma}
\end{subtable}
\caption{\textbf{(a)} Ablation study comparing outer loss functions in multi-step AT with CE as the inner maximization loss. Evaluated using PGD-20 and AutoAttack (AA). Both best and last checkpoint results are reported. \textbf{(b)} Ablation study of hyperparameters $\boldsymbol{\gamma}$ and $\boldsymbol{\beta}$ on model performance in multi-step AT with Resnet-18 on CIFAR-10, evaluated using APGD-CE.}
\label{tab:combined_results}
\end{table*}

\begin{figure*}[ht]
\centering
\resizebox{\textwidth}{!}{
\begin{tabular}{l*{12}{c}}
\toprule
& \multicolumn{6}{c}{\textbf{Hybrid models}} & \multicolumn{2}{c}{\textbf{Robust classifiers}} & \multicolumn{2}{c}{\tbf{Ours \textnormal{\textit{(Global init)}}}~\textbf{\cite{mirza2024shedding}}} & \multicolumn{2}{c}{\tbf{Ours \textnormal{\textit{(Local init)}}}} \\ 
\cmidrule(lr){2-7} \cmidrule(lr){8-9} \cmidrule(lr){10-11} \cmidrule(lr){12-13}
\textbf{Metric} 
& JEM & DRL & JEAT & JEM++ & SADA-JEM & M-EBM 
& PreJEAT & SAT 
& SAT & Better DM 
& SAT & Better DM \\
& \cite{grathwohl2019your} & \cite{gao2020learning} & \cite{zhu2021towards} & \cite{yang2021jem++} & \cite{yang2023towards} & \cite{yang2023mebm} 
& \cite{zhu2021towards} & \cite{santurkar2019singlerobust} 
& \cite{santurkar2019singlerobust} & \cite{wang2023better} 
& \cite{santurkar2019singlerobust} & \cite{wang2023better} \\
\midrule
FID $\downarrow$ & 38.4 & \textbf{9.60} & 38.24 & 37.1 & 9.41 & 21.1 & 56.85 & --- & 45.19 & 30.74 & 57.48 & 49.79 \\
IS $\uparrow$ & 8.76 & 8.58 & 8.80 & 8.29 & 8.77 & 7.20 & 7.91 & 7.5 & 8.58 & \textbf{8.97} & 8.39 & 8.63 \\
\bottomrule
\end{tabular}
}
\captionsetup[table]{width=0.99\linewidth}
    \captionof{table}{Ablation study for different components of our framework using only robust classifiers. Adoption of $\ell_2$ leads to major improvements in metrics. The model from~\cite{wang2023better} outperforms SOTA generative abilities.}
    \label{tab:sota-gen}
\end{figure*}

\minisection{Reweighting with fixed weights}
We suspect that a part of the improved robustness from reweighting may arise \emph{solely} from down-weighting correctly classified samples. At the same time, we hypothesize that some of the gains observed in methods that scale the entire loss term—such as MAIL-AT~\cite{liu2021probabilistic} and WEAT~\cite{mirza2024shedding}—may instead result from scaling the loss by a factor less than one, which indirectly affects the weight decay, a factor known to impact robustness. To isolate these effects, we compare normalized and unnormalized versions of the weighting scheme to assess how much such scaling contributes to the differences. In these experiments, shown in~\cref{tab:ablation_multi_step}, the weighting factor \( w \) is fixed: $w = 10^{-5}$ for correctly classified samples and \( w = 0.1 \) for incorrectly classified samples. We see the normalization yields improvements in CIFAR-10, primarily due to down-weighting correct samples, but brings no gain in CIFAR-100.  In contrast, we find the unnormalized version leads to improvements in both datasets, especially CIFAR-100—likely due to the implicit effect on weight decay. However, in both schemes we observe that the improvements come at the cost of significantly reduced robustness under AA. We believe this drop in AA—also evident in MAIL-AT—stems from down-weighting correctly classified samples, thereby reducing their contribution to the loss and making it more vulnerable to targeted adversarial perturbations. While a detailed investigation of this behavior is beyond the scope of this work, we adopt \textbf{normalized weighting} throughout this paper for methods such as WEAT and MAIL-AT to ensure that observed improvements can be attributed solely to the reweighting of the training samples.

\subsection{Results on Generation with Robust Classifiers} 
Our approach builds on~\cite{mirza2024shedding}, targeting the trade-off between image variance and quality to improve data diversity with a slight compromise in visual fidelity. It reduces data redundancy during generation, producing more diverse outputs compared to~\cite{mirza2024shedding}, as shown in~\cref{fig:qualitative-comparison}.
By initializing from principal components derived from a localized region of the data manifold, the initial representation captures the most salient features of that area. This constrains the optimization process to a specific subspace, resulting in solutions that are more localized and better aligned with the characteristics of that region. Furthermore, since each cluster is created from a different starting image, distinct areas of the data manifold are explored for each case, thereby promoting overall data variability. We conduct experiments on synthetic image generation, with results presented in \cref{tab:sota-gen}. Our findings confirm that the initialization strategy proposed in~\cite{mirza2024shedding} continues to enhance generation performance, even when the initial cluster does not encompass the entire dataset. However, the results also indicate a trade-off between generation quality and data variability. Specifically, the variability introduced by using smaller, distinct clusters leads to a slight decrease in performance, primarily reflected in the FID scores for both BetterDM~\cite{wang2023better} and SAT~\cite{madry2017towards}, as well as in the IS. The performance achieved by our method remains consistent with~\cite{mirza2024shedding}, reaching IS values and FID scores that are comparable to the models that not explicitly trained for generation, while simultaneously increasing data diversity.

\section{Conclusions}\label{sec:conclusions}

Most recent works in AT focus on developing methods to surpass SOTA adversarial robustness, sometimes by adding extra data, augmentations, or parameter-correction techniques like AWP on top of their approach. We instead want to understand the root causes of overfitting in AT using an energy-based framework and improve robustness with minimal modifications, ensuring that the observed gains can be directly attributed to the changes we introduce. By analyzing how the energy dynamics change when perturbing an original sample, we reveal that variations in energy, specifically, the differences in marginal and joint energies, serve as key indicators of both RO and CO. Building on this insight, we introduce a novel regularizer, DER, which promotes a smoother energy landscape during training and effectively mitigates  both RO and CO. We further provide a comprehensive analysis of the role of energy in leveraging the intrinsic generative capabilities of robust classifiers, establishing it as a key reference signal for sample generation. Building on prior work, we propose a training-free framework that enables the generation of synthetic data samples that match the SOTA performance and also mitigate the bias issues observed in previous approaches.
 
\bibliographystyle{IEEEtran}
\bibliography{model_inversion}
\begin{IEEEbiography}[{\includegraphics[width=1in,height=1.25in,clip,keepaspectratio]{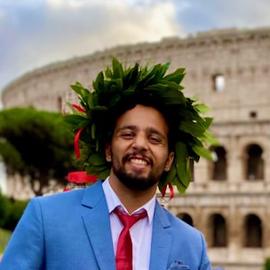}}]{Mujtaba Mirza Hussain} is a PhD student in Computer Science at Sapienza University of Rome. He holds a Bachelor's degree from the University of Kashmir in Computer Science Engineering and a Master's degree from Sapienza. His research interests include making AI systems safer and more trustworthy, with a focus on adversarial machine learning. 
\end{IEEEbiography}
\vskip -2\baselineskip plus -1fil
\begin{IEEEbiography}[{\includegraphics[width=1in,height=1.25in,clip,keepaspectratio]{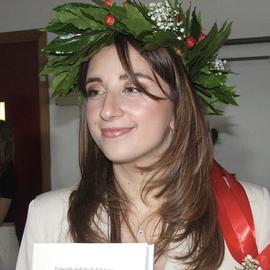}}]{Maria Rosaria Briglia}  is a second-year Ph.D. student in AI Security at Sapienza University as part of the National PhD in Artificial Intelligence. Her research focuses on studying the fundamental behaviors of generative models under adversarial training frameworks and inverse problems, with a particular emphasis on diffusion models. She specializes in enhancing AI robustness, security, and trustworthiness through adversarial methodologies. 
\end{IEEEbiography}
\vskip -2\baselineskip plus -1fil
\begin{IEEEbiography}[{\includegraphics[width=1in,height=1.25in,clip,keepaspectratio]{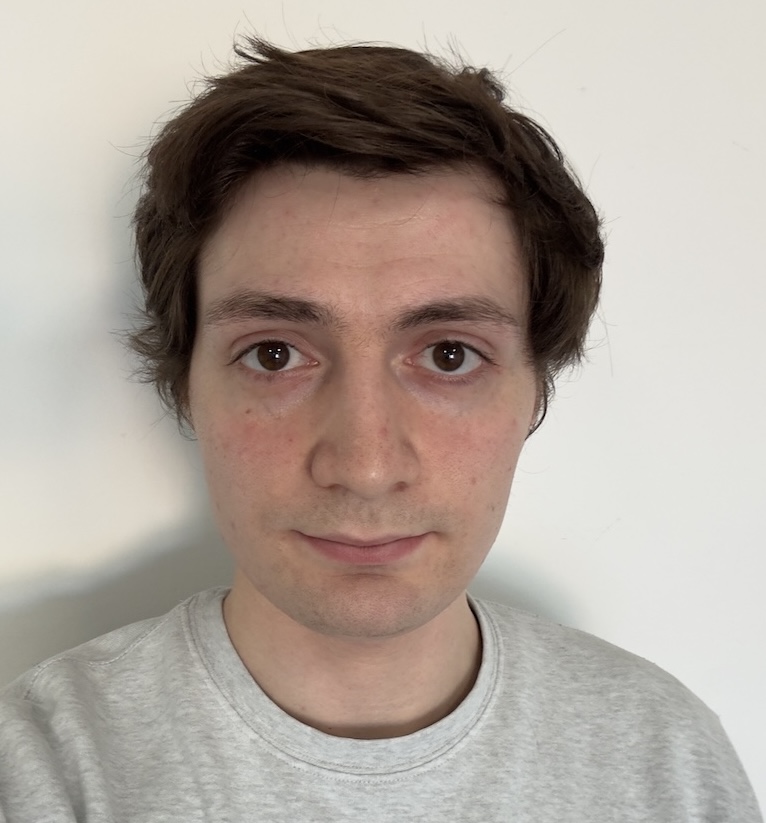}}]{Filippo Bartolucci} is currently a research fellow at the University of Bologna. His current research activity focuses on Artificial Intelligence and Deep Learning techniques for Computer Vision.
\end{IEEEbiography}
\vskip -2\baselineskip plus -1fil
\begin{IEEEbiography}[{\includegraphics[width=1in,height=1.25in,clip,keepaspectratio]{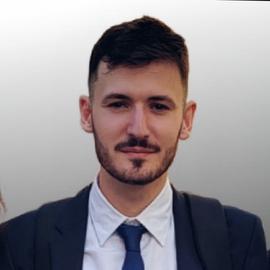}}]{Senad Beadini} is currently an AI Research Engineer at Eustema S.p.A after graduating with a master's degree in Computer Science from Sapienza University of Rome. He is specialized in Adversarial Machine Learning, LLM and NLP.
\end{IEEEbiography}
\vskip -2\baselineskip plus -1fil
\begin{IEEEbiography}[{\includegraphics[width=1in,height=1.25in,clip,keepaspectratio]{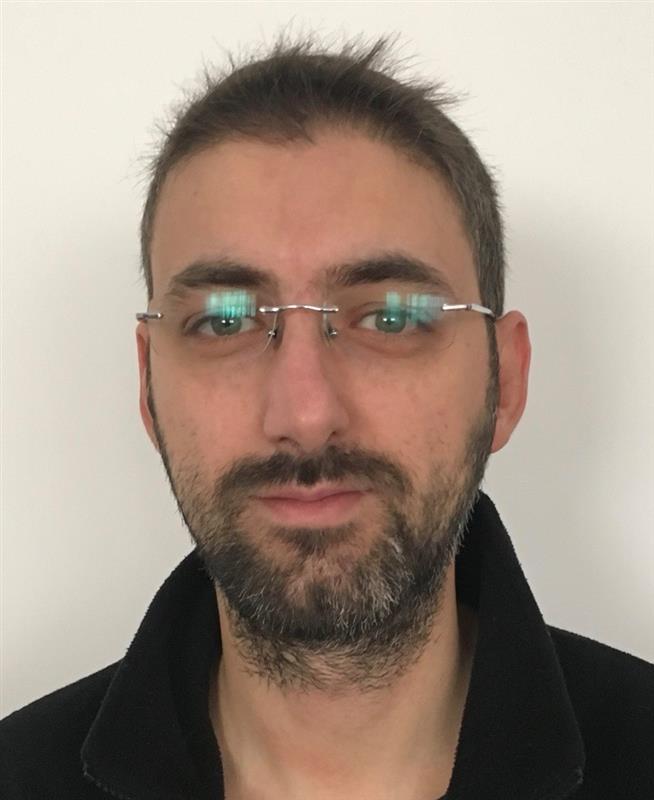}}]{Giuseppe Lisanti} is currently an Associate Professor in the Department of Computer Science and Engineering at the University of Bologna. He has co-authored over 50 publications, with his primary research interests focused on computer vision and the application of deep learning to computer vision problems.
He actively collaborates with other research centers and has participated in various roles in multiple research projects. In 2017, he received the Best Paper Award from the IEEE Computer Society Workshop on Biometrics.
\end{IEEEbiography}
\vskip -1\baselineskip plus -1fil
\begin{IEEEbiography}[{\includegraphics[width=1in,height=1.25in,clip,keepaspectratio]{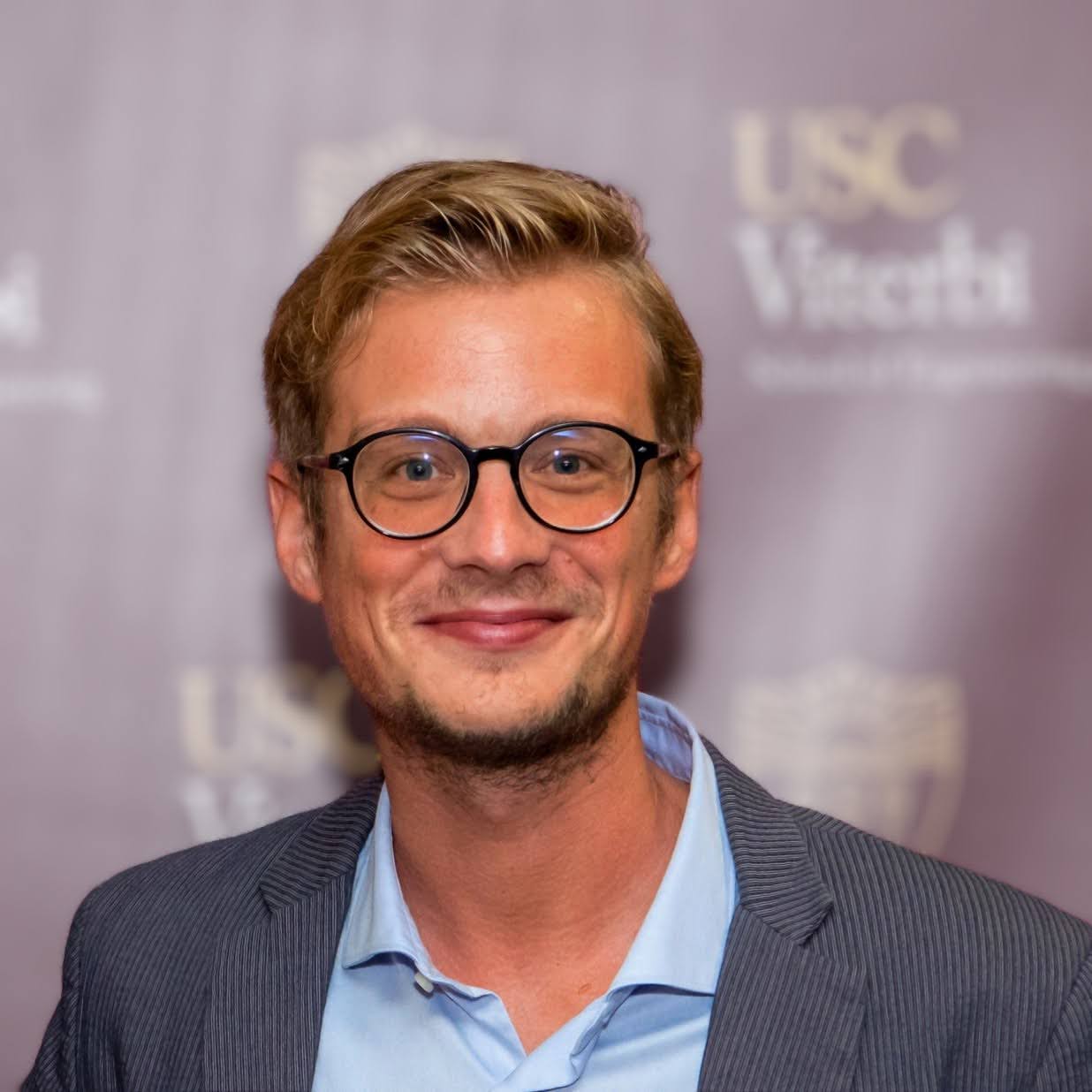}}]{Iacopo Masi}is an Associate Professor in the Computer Science Department at Sapienza, University of Rome, PI and founder of the OmnAI Lab. Till August 2022, he was also Adjunct Research Assistant Professor in the Computer Science Department at the University of Southern California (USC). Previously, Masi was Research Assistant Professor and Research Computer Scientist at the USC Information Sciences Institute (ISI). Masi has been Area-Chair of several conferences in computer vision (WACVs, ICCV’21, ECCV’22, CVPR'24). Masi was awarded the prestigious Rita Levi Montalcini award by the Italian government in 2018. His background covers topics such as tracking, person re-identification, 2D/3D face recognition, adversarial robustness, and facial manipulation detection.
\end{IEEEbiography}
\end{document}